\documentclass[10pt,twocolumn,letterpaper]{article}

\usepackage[pagenumbers]{cvpr} % To force page numbers, e.g. for an arXiv version

\usepackage[OT1]{fontenc} 
\usepackage{graphicx}
\usepackage{amsmath} 
\usepackage{amssymb}
\usepackage{booktabs}
\usepackage[hyphens]{url}  % 
\usepackage{amsfonts}       % blackboard math symbols
\usepackage{nicefrac}       % compact symbols for 1/2, etc.
\usepackage{microtype}      % microtypography
\usepackage{xcolor}         % colors
\usepackage{bibentry}
\usepackage[switch]{lineno}

\usepackage{graphicx}
\usepackage{amsmath}
\usepackage{amsthm}
\usepackage{enumitem}
\usepackage{wrapfig}
\usepackage{caption}
\usepackage{subcaption}
\usepackage{multirow}
\usepackage{pifont}
\usepackage{wasysym}
\usepackage{arydshln}
\usepackage[ruled]{algorithm2e}
\newcommand\like[1]{\begin{picture}(1,1)
	\ifnum0=#1\put(.5,.35){\circle{1}}\else
	\ifnum10=#1\put(.5,.35){\circle*{1}}\else
	\put(.5,.35){\circle{1}}\put(.5,.35){\circle*{.#1}}
	\fi\fi\end{picture}}
\newcommand{\cmark}{\ding{51}}%
\newcommand{\xmark}{\ding{55}}%

\DeclareMathOperator*{\argmax}{arg\,max}

\def\bw{\mathbf{w}}
\def\bx{\mathbf{x}}
\def\bu{\mathbf{u}}
\def\br{\mathbf{r}}
\def\DD{\mathcal{D}} 
\def\UU{\mathcal{U}}
\def\DL{\mathcal{L}}
\def\MS{\mathcal{S}}
\def\AA{\mathcal{A}}

\theoremstyle{definition}
\newtheorem{definition}{Definition}

\usepackage[pagebackref,breaklinks,colorlinks]{hyperref}

\usepackage[capitalize]{cleveref}
\crefname{section}{Sec.}{Secs.}
\Crefname{section}{Section}{Sections}
\Crefname{table}{Table}{Tables}
\crefname{table}{Tab.}{Tabs.}
\crefname{algorithm}{Alg.}{Alg.}

\def\cvprPaperID{6105} % *** Enter the CVPR Paper ID here
\def\confName{CVPR}
\def\confYear{2023}

\begin{document}

\title{Federated Semi-Supervised Learning with Annotation Heterogeneity}
\author{%
	Xinyi Shang\thanks{Equal Contribution} $^{\rm 1}$ \quad Gang Huang$^{\ast\rm 2}$ \quad Yang Lu\thanks{Corresponding author: Yang Lu, luyang@xmu.edu.cn} $^{\rm 1}$ \quad Jian Lou$^{\rm 3}$\quad
	Bo Han$^{\rm 4}$\quad
	Yiu-ming Cheung$^{\rm 4}$\quad
	Hanzi Wang$^{\rm 1}$\\
	$^{\rm 1}$Xiamen University\quad
	$^{\rm 2}$Zhejiang Lab\quad
	$^{\rm 3}$Zhejiang University\quad
	$^{\rm 4}$Hong Kong Baptist University\\
}

% \author{Xinyi Shang\\Xiamen University
% \\{\tt\small shangxinyi@stu.xmu.edu.cn}
% }\and
% \author{Xinyi Shang\\Xiamen University
% \\{\tt\small shangxinyi@stu.xmu.edu.cn}

\maketitle
%%%%%%%%% ABSTRACT
\begin{abstract}
  Federated Semi-Supervised Learning (FSSL) aims to learn a global model from different clients in an environment with both labeled and unlabeled data. Most of the existing FSSL work generally assumes that both types of data are available on each client. In this paper, we study a more general problem setup of FSSL with annotation heterogeneity, where each client can hold an arbitrary percentage (0\%{\text -}100\%) of labeled data. To this end, we propose a novel FSSL framework called Heterogeneously Annotated Semi-Supervised LEarning (HASSLE). Specifically, it is a dual-model framework with two models trained separately on labeled and unlabeled data such that it can be simply applied to a client with an arbitrary labeling percentage. Furthermore, a mutual learning strategy called Supervised-Unsupervised Mutual Alignment (SUMA) is proposed for the dual models within HASSLE with global residual alignment and model proximity alignment. Subsequently, the dual models can implicitly learn from both types of data across different clients, although each dual model is only trained locally on a single type of data. Experiments verify that the dual models in HASSLE learned by SUMA can mutually learn from each other, thereby effectively utilizing the information of both types of data across different clients. 
\end{abstract}

%%%%%%%%% BODY TEXT
\section{Introduction}
Federated Learning (FL) is a machine learning paradigm to learn from data distributed on different data holders in a privacy-preserving manner \cite{mcmahan2017communication,hard2018federated,yang2019federated,2020modelfusion,kairouz2021advances}. %\cite{2015Privacy,mcmahan2017communication,2017multitask,hard2018federated,yang2019federated,2020modelfusion,kairouz2021advances,LiangPanMing2021}.
In FL, locally trained models are transmitted instead of the raw data.
The first FL method FedAvg \cite{mcmahan2017communication} and its following works \cite{li2021fedbn,li2021federated,huang2021personalized} mainly consider supervised learning scenario and only the labeled data on each client is utilized.
In addition to supervised learning, Semi-Supervised Learning (SSL) \cite{lee2013pseudo} has shown its power in various applications.
When both labeled and unlabeled data are available, SSL methods can exploit the information of unlabeled data for a model with better generalization ability.
In the FL environment, a more common scenario is that there are also unlabeled data available on clients, which is termed Federated Semi-Supervised Learning (FSSL) \cite{jeong2020federated}. 
\begin{figure}[!t] 
	\centering
	\includegraphics[width=\linewidth]{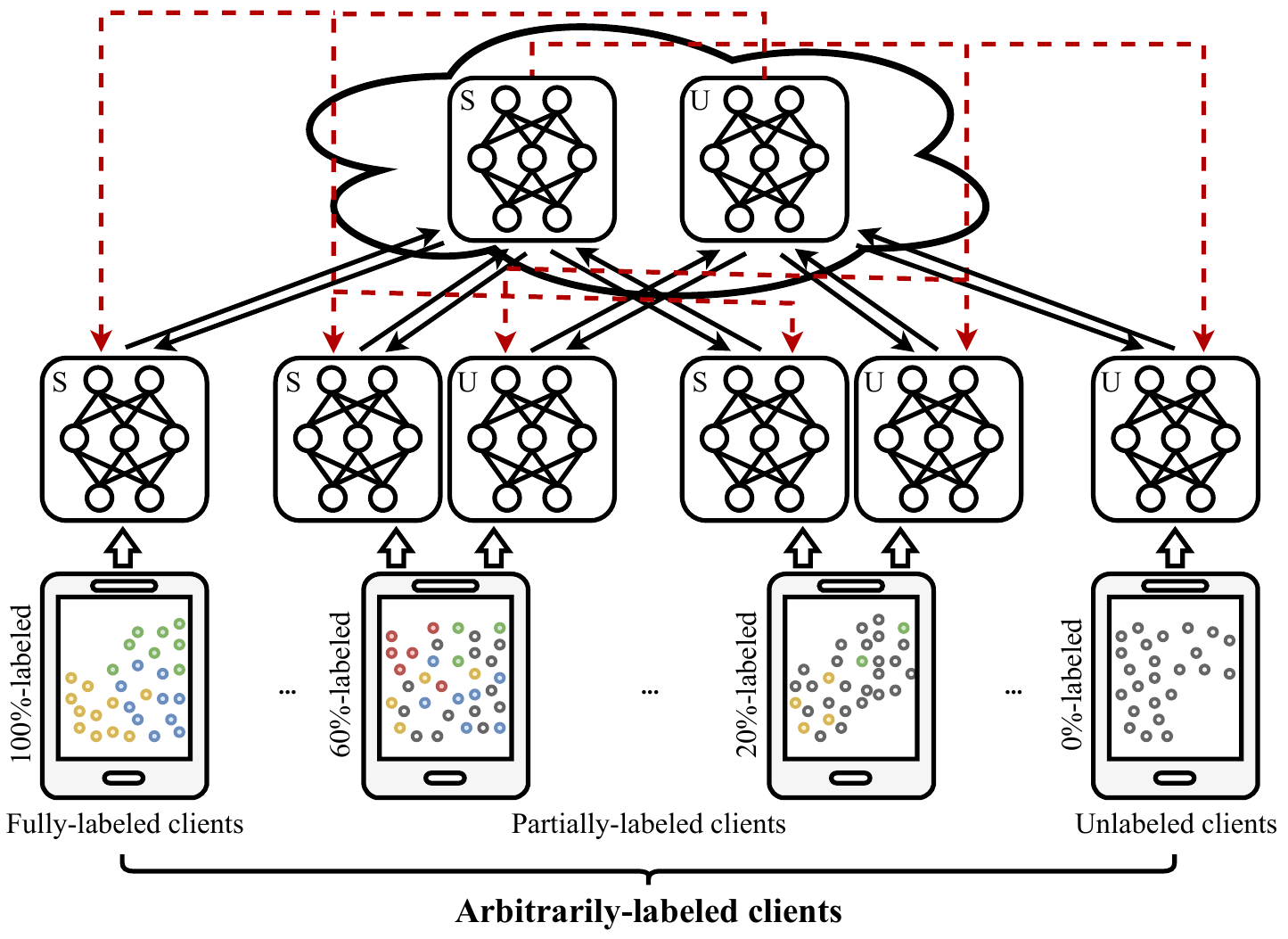}
	\caption{The FL framework with annotation heterogeneity. Colored circles mean labeled data, and grey circles mean unlabeled data. To address the problem of annotation heterogeneity, the proposed HASSLE adopts the dual models (marked as `S' and `U' for `Supervised' and `Unsupervised', respectively) to learn from labeled and unlabeled data separately. The dual models are designed to be mutually improved by the knowledge transfer from each other (red arrows).}
	\label{motivation}
	\vspace{-10pt}
\end{figure}

%联邦下半监督的难度，正式定义annotation heterogeneity
The common assumption of FSSL is that each client holds both labeled and unlabeled data, and the percentage of labeled data is the \emph{same} across all clients \cite{jeong2020federated,DBLP:journals/corr/abs-2008-06180}. However, this assumption is limited to a wide range of applications as not every client has both types of data. From the practical perspective, each client's annotation ability (e.g., availability of experts or annotation budget) should be different. Some clients may have all possessed data labeled, while the others may only have a portion of data labeled or even no data labeled. %\cite{mcmahan2017communication,2017multitask,hard2018federated,2020modelfusion} \cite{2021FederatedCOVID,lu2022federated,jeong2020federated}.
% 注意，所有的claim都需要citation做支持
For example, an FL system is designed to learn a model for the diagnosis of medical records. The experienced diagnosticians
% who can annotate the training data accurately 
are only distributed in a few hospitals \cite{liu2021federated}, which makes the annotation ability of each hospital different. 
In this case, the clients with different percentages of labeled data and the clients with totally unlabeled data impose additional restrictions on the learning process of the existing FSSL methods.
%\color{red}
Therefore, in the paper, we formally defines this more general and practical problem as FSSL with \emph{annotation heterogeneity} (see \cref{definition}), where an FL global model is desired to be trained on \emph{arbitrarily-labeled clients} including fully-labeled, partially-labeled and unlabeled clients. 
%%%%改！
%This problem presents a more general FSSL scenario.
% and also poses new challenges to FL model training.
\cref{motivation} illustrates the problem with annotation heterogeneity.
%, where data heterogeneity \cite{li2019convergence} is also included. 
%i.e., each client holds different classes of labeled data, is also included. 

%% 为什么现存方法不行
\begin{figure}
	\centering
	\includegraphics[width=0.4\textwidth]{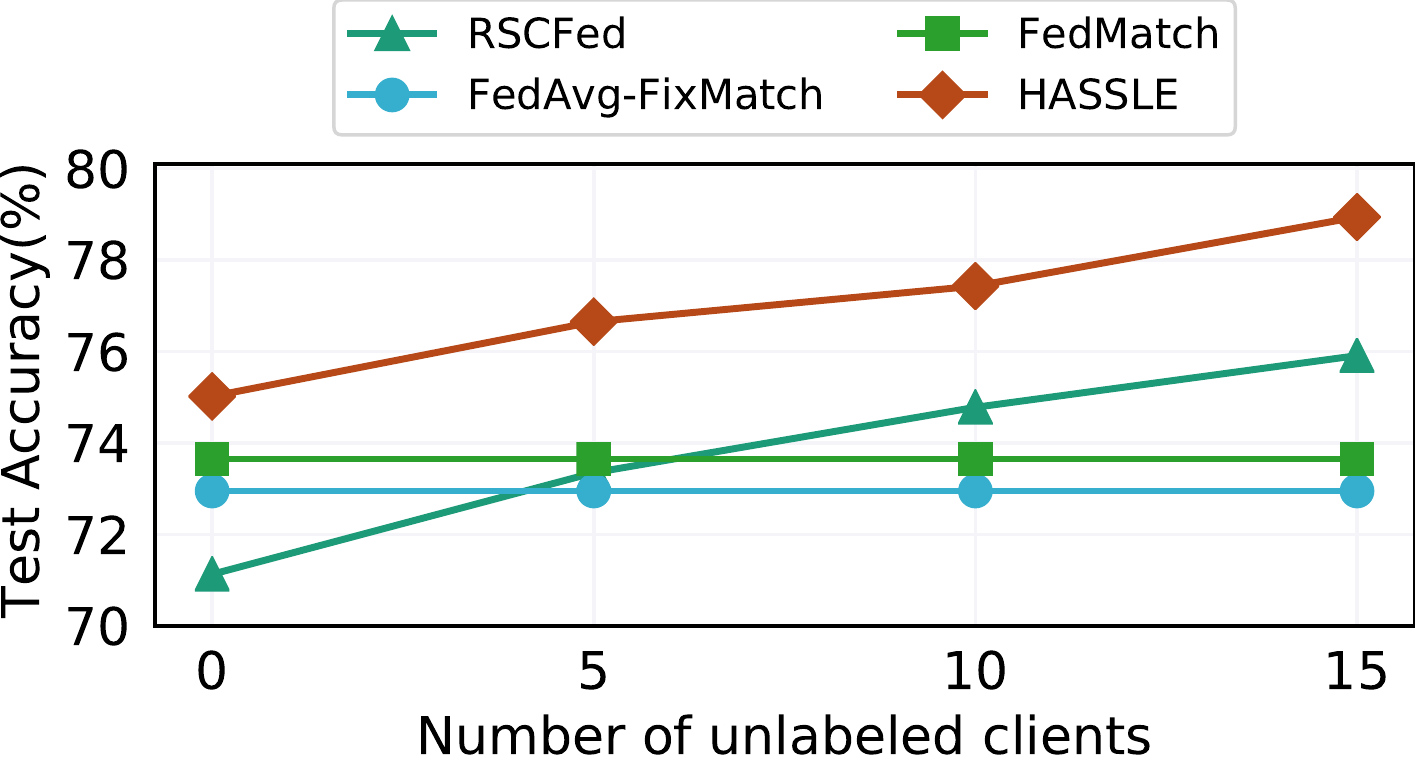}
	\caption{Performance of different FSSL methods with different numbers of participated unlabeled clients. There are five partially-labeled clients for all cases.}
	\vspace{-10pt}
	\label{motivation_exp}
\end{figure}

%\color{red}
% 一般方法做不到，我们能做到
With annotation heterogeneity, most of the existing methods are unable to learn from arbitrarily-labeled clients. On the one hand, typical supervised FL methods \cite{zhao2018federated,li2019convergence,MLSYS2020_38af8613,NEURIPS2020_18df51b9,karimireddy2020scaffold,huang2021personalized} can only learn from clients with labeled data. The knowledge of abundant unlabeled data is wasteful.
% to these supervised FL methods. 
On the other hand, whether existing FSSL methods \cite{zhou2019collaborative,jeong2020federated,2021SemiFed,che2021fedtrinet,wang2021federated,DBLP:journals/corr/abs-2008-06180} or local SSL methods with FL aggregation can only utilize the unlabeled data on the clients that also contain labeled data, i.e., partially-labeled clients. They cannot be directly applied to cpmpletely unlabeled clients. 
%%改！把semifl加上
% clients have completely unlabeled data and can train multiple local epochs to reduce communication costs, while the server has a small amount of labeled data. 
Some other FSSL works \cite{2021FederatedCOVID,liang2022rscfed,diao2021semifl} study some special cases of FSSL. For example, \cite{2021FederatedCOVID,liang2022rscfed} address the case where only fully-labeled clients and unlabeled clients are presented in the FL environment, but the partially-labeled clients are not considered. SemiFL\cite{diao2021semifl} considers the scenario where clients have completely unlabeled data, and the server has a small amount of labeled data. However, all of them lack of consideration of arbitrarily-labeled clients.
\color{black}
To demonstrate the influence of unlabeled clients in FSSL, we conduct an experiment on CIFAR-10 with different numbers of unlabeled clients, as illustrated in \cref{motivation_exp}.
It can be observed that as the number of participated unlabeled clients increases, the performance of FedAvg+FixMatch and FedMatch is not improved because they are not applicable to unlabeled clients. RSCFed takes unlabeled clients into account, but it does not fully exploit the information of partially-labeled clients. Therefore, the key challenge in dealing with annotation heterogeneity is how to effectively extract the knowledge of arbitrarily-labeled clients by model transmission and aggregation with the least influence of the heterogeneous annotation of different clients.

To address the problem of annotation heterogeneity, we propose a novel FL framework called Heterogeneously Annotated Semi-Supervised LEarning (HASSLE) in this paper. Specifically, it is a dual-model framework, where one model is only trained on labeled data, and the other is only trained on unlabeled data. In this manner, the dual models can be easily applied to any client regardless of its labeling percentage.
% During local training, the dual models mutually learn from each other by transferring the knowledge between labeled and unlabeled data across different clients, as shown in Figure \ref{motivation}. 
Furthermore, we propose a mutual learning strategy for HASSLE called Supervised-Unsupervised Mutual Alignment (SUMA), as shown in \cref{motivation}, which makes the dual models mutually learn from each other by transferring the knowledge between labeled and unlabeled data across different clients.
As shown in \cref{motivation_exp}, the performance of the proposed method increases with more unlabeled clients participated, which verifies that our method can well utilize the unlabeled data whether it is on partially-labeled or unlabeled clients.
%\color{red}
In summary, the main contributions of this paper are summarized as follows: 
\begin{itemize}[leftmargin=*]
	\item We define a more realistic and general problem of FSSL with \emph{annotation heterogeneity} (see \cref{definition}), where the percentage of labeled data is arbitrary across clients.
	\item We propose a novel dual-model FL framework HASSLE in \cref{dual_model_framework} to address the problem of annotation heterogeneity, which can be simply applied to any kind of client. 
	\item We propose a mutual learning strategy called SUMA for HASSLE in \cref{mutual_learning}. It enables the dual models to learn from both types of data across different clients.
	\item Experiments show that the proposed HASSLE %significantly outperforms the state-of-the-art FSSL methods and FedAvg+SSL methods with annotation heterogeneity. %\color{red}. It 
	achieves significant accuracy improvements by 2.5\%, 4.6\% and 5.3\% on FMNIST and CIFAR-10/100 compared with the state-of-the-art FSSL method when only 5\% of labeled data is available with annotation heterogeneity.
\end{itemize}
\color{black}
\vspace{-5pt}
\section{Overview of Related Work}
\subsection{Semi-Supervised Learning}
We make an overview of the existing SSL methods. A straightforward way is to assign pseudo-labels on unlabeled data and then involve them in supervised learning \cite{lee2013pseudo,arazo2020pseudo}. 
%These methods fall into the same category of self-training methods \cite{rosenberg2005semi} or entropy minimization-based methods \cite{grandvalet2004semi}.
Many methods are based on the consistency loss, which forces the model to generate consistent outputs when its input is perturbed, such as 
%the ladder network \cite{rasmus2015semi}, $\Pi$ model \cite{laine2016temporal}, 
mean teacher \cite{NIPS2017_68053af2}, virtual adversarial training \cite{miyato2018virtual}, and UDA \cite{xie2020unsupervised}.
%such as the ladder network \cite{rasmus2015semi}, $\Pi$ model \cite{laine2016temporal}, mean teacher \cite{NIPS2017_68053af2}, virtual adversarial training \cite{miyato2018virtual}, and UDA \cite{xie2020unsupervised}.
Recently, SSL methods that combine consistency of augmented data and label generation have achieved the promising performance, such as MixMatch \cite{berthelot2019mixmatch}, ReMixMatch \cite{berthelot2019remixmatch}, FixMatch \cite{sohn2020fixmatch} and SimPLE \cite{hu2021simple}.
It is worthwhile to point out that there are unlabeled clients under annotation heterogeneity, which makes a naive combination of the centralized SSL techniques and FL algorithms hardly achieve satisfactory performance.
%%%%%还可以强调有标签和无标签的分布不同
%\color{red}
\subsection{Federated Semi-Supervised Learning}
%With annotation heterogeneity, most of the existing methods are limited to learning from arbitrarily-labeled clients.
%The knowledge of abundant unlabeled data is wasteful to these supervised FL methods.
%On the other hand, existing FSSL methods \cite{jeong2020federated,wang2021federated}, as well as locally applying SSL methods with global aggregation, can only utilize the unlabeled data on the clients that also contain labeled data, i.e., partially-labeled clients.  
%They cannot be directly applied to unlabeled clients.
%Some other FSSL work \cite{2021FederatedCOVID,liang2022rscfed} studies a special case of FSSL, where only fully-labeled clients and unlabeled clients are presented in the FL environment, and the partially-labeled clients are not considered.

%FSSL studies the case that not all data is labeled in the FL environment.
% 很多方法都是假设分布相同
The common assumption of FSSL is that each client owns both labeled and unlabeled data \cite{jeong2020federated,2021SemiFed,che2021fedtrinet,wang2021federated,liu2021federated,DBLP:journals/corr/abs-2008-06180}. 
%For example, FedMatch \cite{jeong2020federated} is proposed with a new inter-client consistency loss that regularizes the models learned at local clients to output the same prediction. 
However, they can only utilize the unlabeled data on the clients that also contain labeled data, i.e., partially-labeled clients, and cannot be directly applied to unlabeled clients. Some other FSSL works \cite{2021FederatedCOVID,liang2022rscfed,diao2021semifl} study some special cases of FSSL. For example, \cite{2021FederatedCOVID,liang2022rscfed} study the scenario where only fully-labeled clients and unlabeled clients are presented in the FL environment, but the partially-labeled clients are not considered. 
%SemiFL\cite{diao2021semifl} considers the case where clients have completely unlabeled data, and the server has a small amount of labeled data. %However, all of them do not consider arbitrarily-labeled clients.
%SemiFed \cite{2021SemiFed} unifies consistency regularization and pseudo-labeling for the FL environment. 
% 有两个special case
%Moreover, \cite{wang2021federated} considers the class distribution mismatch between labeled and unlabeled data and proposes two regularization terms to alleviate this mismatch in FSSL.
%%%我们工作的优势
%However, these methods do not consider arbitrarily-labeled clients.
%Some other FSSL works \cite{2021FederatedCOVID,liu2021federated,liang2022rscfed} study some special cases of FSSL, where only fully-labeled and unlabeled clients are presented in the FL environment, and the partially-labeled clients are not considered.
Unlike the work mentioned above, our work is the first attempt to consider and learn from arbitrarily-labeled clients effectively. 
\color{black}
%Some other FSSL works \cite{2021FederatedCOVID,liu2021federated,liang2022rscfed} study a special case of FSSL, where only fully-labeled and unlabeled clients are presented in the FL environment, and the partially-labeled clients are not considered.
%Unlike the work mentioned above, our work is the first attempt to consider arbitrarily-labeled clients. 
%To address the problem, we propose a novel FL framework HASSLE with a specific mutual learning strategy SUMA.
%%
\section{Preliminary and Problem Formulation}
\vspace{-3pt}
%%怎么去定义annotation heterogeneity
\paragraph{Federated Learning} 
A typical FL environment includes a server and $K$ clients. 
%Each client has a local dataset $\DD_k$ and a corresponding local model $\bw_k^{(t)}$ in round $t$. 
In round $t$, the server first sends a global model $\bw^t$ to clients. The clients then update the received model on their local data $\DD^k$. Then, some clients are selected to upload their updated models to the server. Finally, the server performs weighted average to update the global model for round $t+1$.
%\color{red} Each client takes the received global model $\bw^{(t)}$ as the initialized local model and continually updates it locally on $\DD_k$ to obtain $\bw_k^{(t+1)}$.\color{black}
\vspace{-10pt}
\paragraph{Annotation Heterogeneity}
The problem of FSSL is how to utilize the unlabeled data on clients properly. 
Most of the existing FSSL methods assume that unlabeled data is distributed on every client \cite{jeong2020federated}.
However, just like the distribution difference of each client, the annotation ability of each client may also be different, which means that not every client owns both types of data.
Motivated by this more realistic scenario, we consider a new FSSL problem called annotation heterogeneity, formally defined below.
\vspace{-1pt}
\begin{definition}[\emph{Annotation Heterogeneity}]\label{definition}
	\emph{Suppose only a fraction of data is labeled in federated learning, annotation heterogeneity is the problem that the percentage of labeled data is arbitrary across clients.}
\end{definition}

With annotation heterogeneity, we call each client an \emph{arbitrarily-labeled client} because each client can be either a fully-labeled client (all data is labeled), an unlabeled client (no data is labeled), or a partially-labeled client ($a\%$ data is labeled, $a\in(0,100)$). 
In this case, the local dataset $\DD_k$ of a partially-labeled client $k$ can be further divided into a labeled subset $\DD_{\DL,k}=\{(\bx_{k,i},y_{k,i})\}_{i=1}^{n_{\DL,k}}$ and an unlabeled subset $\DD_{\UU,k}=\{\bu_{k,i}\}_{i=1}^{n_{\UU,k}}$, where $n_{\DL,k}$ and $n_{\UU,k}$ are the cardinalities of the corresponding subsets.
Similar to the typical SSL setting, the total number of labeled data is far less than the total number of unlabeled data, i.e., $\sum_{k=1}^K\DD_{\DL,k}\ll\sum_{k=1}^K\DD_{\UU,k}$.
The key challenge in dealing with annotation heterogeneity is how to integrate the information from arbitrarily-labeled clients, while most of the existing FSSL methods \cite{jeong2020federated,wang2021federated,DBLP:journals/corr/abs-2008-06180} can only learn from fully-labeled or partially-labeled clients.

\section{Proposed Method}
To specifically address the problem of annotation heterogeneity, we first propose a new dual-model FL learning framework called Heterogeneously Annotated Semi-Supervised LEarning (HASSLE). 
In this framework, we propose a specific mutual learning strategy called SUMA to take full advantage of the dual models.
\vspace{0pt}
\subsection{HASSLE: A Dual-Model Learning Framework}\label{dual_model_framework}
% 动机、挑战、优势
With annotation heterogeneity, the key challenge is how to learn from both labeled and unlabeled data on arbitrarily-labeled clients jointly.
% 有监督半监督在无标签上不行
On the one hand, the most obvious obstacle is that a supervised or semi-supervised FL model cannot learn from the data on an unlabeled client.
% 半监督在标注异构
On the other hand, the performance of an FSSL model is likely to be affected by the annotation percentage difference among partially-labeled clients because the weight of aggregating local semi-supervised models cannot be well measured \cite{liang2022rscfed}.
Therefore, instead of a single model (e.g., a single semi-supervised model) for local training and global aggregation, we propose to use dual models with the same structure to learn from labeled and unlabeled data separately.

% 双模型设定
Formally, we introduce the dual models in HASSLE: a supervised model $\bw_\MS$ is exclusively trained on labeled data, and an unsupervised model $\bw_\UU$ is exclusively trained on unlabeled data.
The update of the dual models follows the typical FL manner in which they are trained locally and aggregated on the server with model transmission.
%\color{red} In round $t$, the local dual models on client $k$ are denoted as $\bw_{\MS,k}^{(t)}$ and $\bw_{\UU,k}^{(t)}$, and the global dual models are denoted as $\bw_{\MS}^{(t)}$ and $\bw_{\UU}^{(t)}$.\color{black}
In round $t$, all updated models, including supervised and unsupervised models, are uploaded to the server and aggregated, respectively:
\begin{align}\label{local_update}
\bw_{\MS}^{(t)} = \frac{1}{\sum_{k\in \AA_\DL^{(t)}} n_{\DL,k}}\sum_{k\in \AA_\DL^{(t)}} n_{\DL,k}\bw_{\MS,k}^{(t)}, \\
\bw_{\UU}^{(t)} = \frac{1}{\sum_{k\in \AA_\UU^{(t)}} n_{\UU,k}}\sum_{k\in \AA_\UU^{(t)}} n_{\UU,k}\bw_{\UU,k}^{(t)},
\end{align}
where $\AA^{(t)}$ is the set of selected clients in round $t$. $\AA_\DL^{(t)}=\{k|k\in\AA^{(t)}, n_{\DL,k}\ne 0\}$ and $\AA_\UU^{(t)}=\{k|k\in\AA^{(t)}, n_{\UU,k}\ne 0\}$ are the subsets of the clients possessing labeled and unlabeled data, respectively.

%% 双模型的优势
There are two main advantages of HASSLE: 1) It can be simply applied to arbitrarily-labeled clients regardless of the client's annotation percentage. 
For partially-labeled clients, both $\bw_{\MS,k}^{(t)}$ and $\bw_{\UU,k}^{(t)}$ can be updated on local subsets $\DD_{\DL,k}$ and $\DD_{\UU,k}$, respectively. 
For fully-labeled clients and unlabeled clients, one of the dual models can be easily updated on the corresponding local data. 2) The aggregation weight in \cref{local_update} only relies on the number of corresponding training samples.
For partially-labeled clients, it is not required to sum up the number of labeled and unlabeled training samples to compute aggregation weight, which naturally solves the problem of weight mismeasurement because the label quality of a labeled sample is usually higher than the quality of the pseudo-label assigned to an unlabeled sample.

%That solves the problem of weight mismeasurement by using a single model because the label quality of a labeled sample is usually higher than the quality of the pseudo-label assigned to an unlabeled sample.

%The aggregation weight in Equation (\ref{local_update}) only relies on the number of corresponding training samples, which is another advantage of using two models. The contribution of labeled and unlabeled data to each model will not be mixed up during aggregation because we usually believe that the value of a labeled sample is higher than an unlabeled sample.
% 关键点在于如何相互学
Now, the key issue of HASSLE is how to make these dual models mutually learn from each other such that the knowledge learned from labeled and unlabeled data can be transferred to both models.
Given \cref{local_update}, the supervised global model contains the information of the labeled data across labeled and partially-labeled clients.
In the meantime, the unsupervised global model contains the information of the unlabeled data across unlabeled and partially-labeled clients.
If the local dual models are separately aggregated on the server, the global dual models only contain limited information.
% and obtain the knowledge of both types of data across different clients.
Therefore, it is necessary to propose a mutual learning strategy by which the unsupervised model may provide more information to the supervised model, and the supervised model may help the unsupervised model better learn from the unlabeled data.
The optimization for the update of local dual models can be formulated by:
\begin{align}\label{hassle_optimization}
\bw_{\MS,k}^{(t+1)}\leftarrow\bw_{\MS}^{(t)} - \eta\nabla L_\MS(\DD_{\DL,k}, \bw_{\MS,k}^{(t)}| \bw_{\UU}^{(t)}),\\
\bw_{\UU,k}^{(t+1)}\leftarrow\bw_{\UU}^{(t)} - \eta\nabla L_\UU(\DD_{\UU,k}, \bw_{\UU,k}^{(t)}| \bw_{\MS}^{(t)}).
\end{align}
When we update $\bw_{\MS,k}^{(t+1)}$ on $\DD_{\DL,k}$, we use the global supervised model $\bw_{\MS}^{(t)}$ as the initial model and design a loss function $L_\MS$ that involves the global unsupervised model $\bw_{\UU}^{(t)}$. Similarly, the update of $\bw_{\UU,k}^{(t+1)}$ on $\DD_{\UU,k}$ also involves $\bw_{\MS}^{(t)}$.
In this manner, the knowledge learned from unlabeled clients can be transferred to the supervised model.
At the same time, the improved supervised model can also better help the unsupervised model.
After mutual learning, both supervised and unsupervised models can be used for inference, as well as their ensemble model that is calculated by the mean outputs of the dual models above.
The comparison of using supervised, unsupervised, and ensemble model for inference is discussed in \cref{section_comparison}.
The dual-model learning framework of HASSLE is illustrated in \cref{motivation}.
%\begin{wrapfigure}{R}{0.3\textwidth}\vspace{-10pt}
%	\includegraphics[width=0.35\textwidth]{figure/method.pdf}  
%	%	\vspace{2pt}
%	\caption{The local training process of SUMA. The number in the circle indicates the operation order.} \vspace{-10pt}
%	\label{training}
%	\vspace{-10pt}
%\end{wrapfigure}

The proposed HASSLE framework resembles the classic Co-Training algorithm \cite{blum1998combining} to some extent because both of them adopt two models for semi-supervised learning. 
However, they are fundamentally different in the following aspects: 1) \textbf{Target task}. Co-Training is designed for batch semi-supervised learning, which cannot be used to solve the problem of annotation heterogeneity in FSSL. 2) \textbf{Training source}. The two models in Co-Training are built on different portions of data, while the two models in HASSLE are separately built on different types of data, i.e., labeled and unlabeled data. 3) \textbf{Knowledge transfer}. Co-Training does not explicitly conduct mutual learning, while HASSLE adopts a specifically designed mutual learning strategy described in \cref{mutual_learning}.

\begin{figure}[!t]
	\centering
	\includegraphics[width=0.45\textwidth]{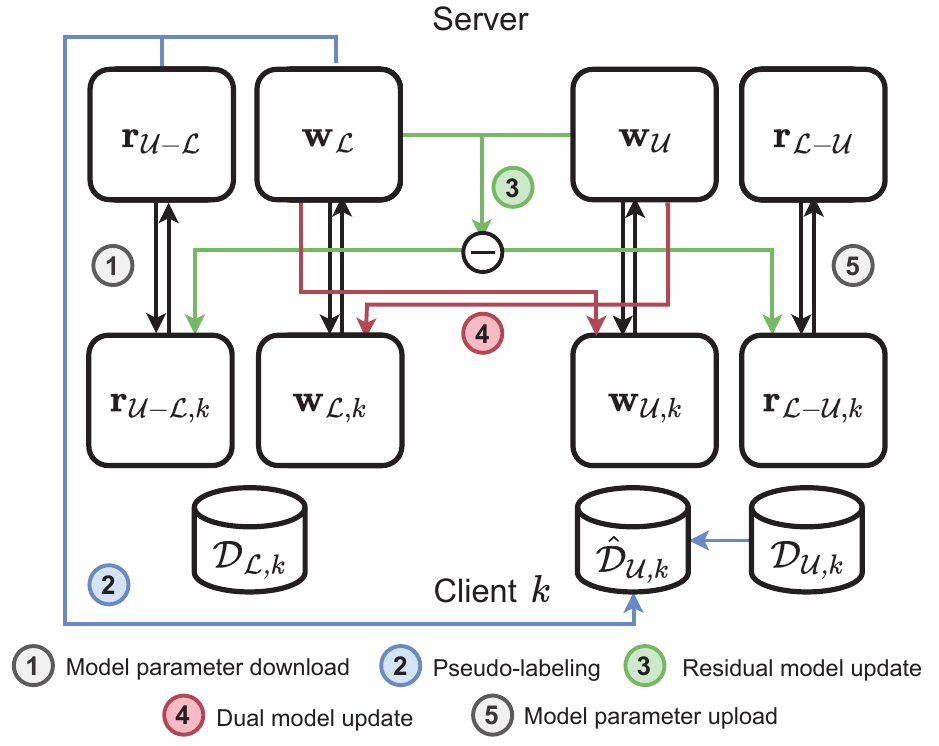}  
	%	\vspace{2pt}
	\caption{The local training process of SUMA in each round. The number in the circle indicates the operation order.} 
	\label{training}
	%\vspace{-10pt}
\end{figure}

\subsection{SUMA: A Mutual Learning Strategy}\label{mutual_learning}
To further improve the dual models with knowledge transfer, we propose a specific mutual learning strategy called Supervised-Unsupervised Mutual Alignment (SUMA). 
In SUMA, the dual models are with the same architecture.
The local unsupervised model is trained in a supervised learning manner on pseudo-labels annotated by the global supervised model. 
In addition to aligning the dual models' learning paradigm, pseudo-labeling is also a straightforward way to transfer the knowledge from the supervised model to the unsupervised model.
Given the global supervised model $\bw_\MS^{(t)}$, the pseudo-labels of the unlabeled data $\DD_{\UU,k}$ are given by:
\begin{align}\label{pseudo_labeling}
\hat{y}^{(t)}_{k,i} = \argmax_{j=1,...,M} f(\bu_{k,i},\bw_\MS^{(t)}),\ \ \ \ i=1,...,n_{\UU,k},
\end{align}
where the function $f(\bx,\bw)$ outputs the logits of the input data $\bx$ with the model $\bw$, and $M$ is the number of classes.
After assigning pseudo-labels by the global supervised model, we obtain local pseudo-labeled subsets $\hat{\DD}_{\UU,k}^{(t)}=\{(\bu_{k,i},\hat{y}_{k,i}^{(t)})\}, i=1,...,n_{\UU,k}$ for all partially-labeled and unlabeled clients.
%\color{red}. In our setting, threshold like FixMatch \cite{sohn2020fixmatch} is not play a major role to select high-confidence unlabaled samples. Although the pseudo-labels may not be accurate in the early stages of training, updating the unsupervised model on unlabeled data is not significantly affected. The main reason is that the proposed model proximity makes the local unsupervised model close to the global supervised model in the parameter space, which guarantees the correct optimization direction to some extent. We compare our method with the case where the threshold is set at 0.95, the same as FixMatch. The results and corresponding analysis are shown in \cref{threshold} in Appendix C.1.

\color{black}
With pseudo-labeling, both the dual models can be learned in a supervised learning manner. Subsequently, we can further improve the dual models from two perspectives of mutual learning.
On the one hand, we adopt two additional compact models called \emph{residual models} to learn the differences between the dual models. On the other hand, we propose to impose an optimization constraint to make the dual models close to each other in the model parameter space during local training.
The training process of SUMA is illustrated in \cref{training}, and the detailed training process and corresponding description for both local clients and the server is shown in \cref{alg:algorithm} in Appendix A.
\vspace{-10pt}
\paragraph{Global Residual Alignment.}
The dual models contain different information because they are trained on different types of data, so the complementary information of the dual models is helpful to each other.
Given a sample $\bx$, the complementary information can be modeled by the difference of logits:
\begin{align}\label{logits_difference}
f(\bx;\bw_\UU^{(t)}) - f(\bx;\bw_\MS^{(t)})\  \textrm{and} \  f(\bx;\bw_\MS^{(t)}) - f(\bx;\bw_\UU^{(t)})
\end{align}
To well utilize the complementary information, we can incorporate two compact models $\br_{\UU-\MS}^{(t)}$ and $\br_{\MS-\UU}^{(t)}$ to learn the difference between $\bw_\UU^{(t)}$ and $\bw_\MS^{(t)}$. 
To be specific, $\br_{\UU-\MS}^{(t)}$ is called the supervised residual model, which aims to learn the knowledge from the global unsupervised model to complement the global supervised model on $\DD_{\DL,k}$.
Similarly, $\br_{\MS-\UU}^{(t)}$ is called the unsupervised residual model to learn from the global supervised model on $\DD_{\UU,k}$. Note that the two residual models are not opposite of each other because they aim to learn totally different knowledge on different types of data.
Here, we only describe how the supervised residual model $\br_{\UU-\MS}^{(t)}$ is updated.
The unsupervised residual model $\br_{\MS-\UU}^{(t)}$ can be updated in the same way. For $\br_{\UU-\MS}^{(t)}$, our goal is to transfer the knowledge of the unsupervised model to make it better.
Therefore, the residual model $\br_{\UU-\MS}^{(t)}$ should learn the knowledge that the unsupervised model owns while the supervised model does not own.

We use two loss functions to update the residual models. The first loss function ensures that the residual models can learn the difference between the dual models.
\begin{equation}\label{res_loss}
\begin{split}
L_{res}  = \sum_{(\bx,y)\in\DD_{\DL,k}}KL\Big(\sigma\big(f(\bx;\br_{\UU-\MS,k}^{(t)})/\tau\big), \\ \sigma\big((f(\bx;\bw_\UU^{(t)})  - f(\bx;\bw_\MS^{(t)}))/\tau\big)\Big),
\end{split}
\end{equation}
where $\tau$ is the temperature hyperparameter, and $\sigma(\cdot)$ is the softmax function. 
%However, the residual models cannot be directly trained in the FL environment.
%They should also follow the process of local training and global aggregation.
%On each client, we first update the local residual models $\br_{\UU-\MS,k}^{(t)}$ and use the Kullback-Leibler (KL) divergence to measure the outputs of the local residual models with the logit differences :
With \cref{res_loss}, the local residual model learns the difference between the global dual models.
However, it is not guaranteed that adding the supervised residual model to the supervised model results in accurate prediction.
Therefore, the second loss function measures how the supervised residual model complements the corresponding supervised model:
\begin{align}\label{res_ce_loss}
L_{CE} = \sum_{(\bx,y)\in\DD_{\DL,k}}\ell\big(f(\bx;\bw_\MS^{(t)}) + f(\bx;\br_{\UU-\MS,k}^{(t)}), y\big),
\end{align}
where $\ell(\cdot,\cdot)$ is the cross-entropy loss.
Then, we can update the local supervised residual model by:
\begin{align}\label{residual_model_update}
\br_{\UU-\MS,k}^{(t+1)} = \br_{\UU-\MS}^{(t)} - \eta\nabla (L_{CE}+\lambda L_{res}),
\end{align}
where $\lambda$ is the trade-off hyperparameter, and its influence is  investigated in \cref{fig_kl_weight} in Appendix C.4, which shows that HASSLE is robust to most of $\lambda$ values. $\br_{\UU-\MS}^{(t)}$ is the global supervised residual model, which is obtained by aggregating local residual models on the server:
\begin{align}\label{residual_model_aggregation}
\br_{\UU-\MS}^{(t)} = \frac{1}{\sum_{k\in \AA_\DL^{(t)}} n_{\DL,k}}\sum_{k\in \AA_\DL^{(t)}} n_{\DL,k}\br_{\UU-\MS,k}^{(t)}.
\end{align}
%The updated local residual models capture the difference between the dual models.
By model aggregation, the global residual models can learn the difference between the dual models from a global perspective, which complements the dual models.

The update of the unsupervised residual model $\br_{\MS-\UU}^{(t)}$ also follows \cref{res_loss}-\cref{residual_model_aggregation} by simply switching the role of supervised and unsupervised model, and changing the local labeled subset $\DD_{\DL,k}$ the local pseudo-labeled subset $\hat{\DD}_{\UU,k}^{(t)}$.

As the residual models are designed to complement the dual models, given a test data $\bx$, the inference is done by $f(\bx,\bw_\MS^{(t-1)}) + f(\bx,\br_{\UU-\MS}^{(t)})$ and $f(\bx,\bw_\UU^{(t-1)}) + f(\bx,\br_{\MS-\UU}^{(t)})$ for supervised and unsupervised model, respectively. 
Note that we use the global supervised model on the previous round here because the global residual model $\br_{\UU-\MS}^{(t)}$ after aggregation is obtained to learn the difference between $\bw_\MS^{(t-1)}$ and $\bw_\UU^{(t-1)}$.
With the dual models, the pseudo-labeling in \cref{pseudo_labeling} is correspondingly changed to:
%\begin{align}\label{pseudo_labeling_with_residual_model}
%\hat{y}^{(t)}_{k,i} = \argmax_{j=1,...,M} \big(f(\bu_{k,i},\bw_\LL^{(t-1)}) + f(\bu_{k,i},\br_{\UU-\LL}^{(t)})\big),\ \ \ \ i=1,...,n_{\UU,k},
%\end{align}
\begin{align}\label{pseudo_labeling_with_residual_model}
\hat{y}^{(t)}_{k,i} = \argmax_{j=1,...,M} \big(f(\bu_{k,i},\bw_\MS^{(t-1)}) + f(\bu_{k,i},\br_{\UU-\MS}^{(t)})\big),
\end{align}

One may notice that ResKD \cite{li2021reskd} also adopts the residual models. However, our targeting problem and method are fundamentally different from ResKD. ResKD iteratively calculates the residuals between the teacher and student model trained on the same data, which aims to approximate the student model to the teacher model with the help of a series of residual models. However, the residual models in our method aim to learn the complementary knowledge between the dual models trained on different types of data. In addition, both the dual models and the residual models follow the manner of FL model training and aggregation. 
%Overall, our method is innovative compared with ResKD.
\vspace{-10pt}
\paragraph{Model Proximity Alignment.} \label{L2}
With the residual models, we can explicitly model the differences between the dual models. However, the dual models themselves are not mutually improved except by transferring knowledge with pseudo-labeling.
In this case, the ability of the supervised model is limited because it is nothing but a FedAvg model on all labeled data.
The unsupervised model is also affected because it relies on the quality of the pseudo-labels given by the supervised model.
If the dual models themselves are not mutually improved, their complementary information may not provide much help to each other.
Consequently, the advantages of the residual models cannot be fully exploited.
Therefore, we propose a model proximity term to align the dual models in the model parameter space when updating them locally.
The update of the local supervised model with proximity alignment is formulated by:
\begin{align}\label{L2_optimization_1}
L_\MS(\DD_{\DL,k}, \bw_{\MS,k}^{(t)}| \bw_{\UU}^{(t)})&=\sum_{(\bx,y)\in\DD_{\DL,k}}\ell\big(f(\bx,\bw_{\MS,k}^{(t)}),y\big),\nonumber\\&+\gamma\|\bw_{\MS,k}^{(t)}-\bw_\UU^{(t)}\|_2,
\end{align}
where $\gamma$ is the trade-off hyperparameter, and its influence is  investigated in \cref{fig_l2_weight} in Appendix C.4, which shows that HASSLE is robust to most of $\gamma$ values.
The update of the local unsupervised model $\bw_{\UU,k}^{(t)}$ also follows \cref{L2_optimization_1} by simply switching the role of supervised and unsupervised model, and changing the local labeled subset $\DD_{\DL,k}$ to the local pseudo-labeled subset $\hat{\DD}_{\UU,k}^{(t)}$.
%\begin{equation}\label{L2_optimization_2}
%\begin{split}
%L_\UU(\DD_{\UU,k}, \bw_{\UU,k}^{(t)}| \bw_{\LL}^{(t)}) = \sum_{(\bu,\hat{y})\in\hat\DD_{\UU,k}}\ell\big(f(\bu,\bw_{\UU,k}^{(t)}),\hat{y}\big) \\+ \gamma\|\bw_\LL^{(t)}-\bw_{\UU,k}^{(t)}\|_2.
%\end{split}
%\end{equation}

%\begin{align}\label{L2_optimization_1}
%L_\LL(\DD_{\LL,k}, \bw_{\LL,k}^{(t)}| \bw_{\UU}^{(t)}) &= \sum_{(\bx,y)\in\DD_{\LL,k}}\ell\big(f(\bx,\bw_{\LL,k}^{(t)}),y\big) + \gamma\|\bw_{\LL,k}^{(t)}-\bw_\UU^{(t)}\|_2,\\\label{L2_optimization_2}
%L_\UU(\DD_{\UU,k}, \bw_{\UU,k}^{(t)}| \bw_{\LL}^{(t)}) &= \sum_{(\bu,\hat{y})\in\hat\DD_{\UU,k}}\ell\big(f(\bu,\bw_{\UU,k}^{(t)}),\hat{y}\big) + \gamma\|\bw_\LL^{(t)}-\bw_{\UU,k}^{(t)}\|_2.
%\end{align}
Using the $L_2$ norm as proximity is similar to the term adopted in \cite{MLSYS2020_38af8613}. However, the targeting problem and the way are different. Previous works adopt $L_2$ norm to regularize local models from the global model, which limits the diversity of the local models to solve the problem of data heterogeneity. However, in SUMA, the motivation of using the $L_2$ norm is primarily for mutual learning, where the local model is required to be close to the other global dual model rather than its corresponding global model.
In addition, using the $L_2$ norm may also inherit the property to deal with data heterogeneity, which is validated by the experimental results in \cref{fig_non_iid} in Appendix C.2.
\vspace{-10pt}
%\color{red}.
\paragraph{Remarks on Threshold for Pseudo-labeling.}
In our setting, threshold like FixMatch \cite{sohn2020fixmatch} does not play a major role in selecting high-confidence unlabaled samples. Although the pseudo-labels may not be accurate in the early stages of training, updating the unsupervised model on unlabeled data is not significantly affected. The main reason is that the proposed model proximity makes the local unsupervised model close to the global supervised model in the parameter space, which guarantees the correct optimization direction to some extent. We compare our method with the case where the threshold is set at 0.95, the same as FixMatch. The results are shown in \cref{threshold} in Appendix C.1.
\color{black}
\vspace{-10pt}
\paragraph{Remarks on Communication Cost.}
Communication cost is an important issue in FL. It is worthwhile to point out that the communication cost of our method is still comparable with other FL methods, although multiple models are updated. On the one hand, the residual models are actually more compact models with only 1/4 channels of the dual models, shown in the fourth part of \cref{model_evaluation}. On the other hand, not all kinds of clients need to transmit four models in the proposed HASSLE framework. For fully-labeled clients and unlabeled clients, each client only needs to upload the models (one dual model and one compact residual model) according to the type of data it holds, leading to a slightly higher communication cost than a single model. Only partially-labeled clients need to update and upload all models, which is almost as same as the communication cost of FedMatch that needs to transmit a double of the model parameters. Nevertheless, transmitting multiple models is worthy because our proposed HASSLE framework can be applied to arbitrarily-labeled clients and effectively extract the knowledge of both labeled and unlabeled, which are beyond the ability of existing FL methods.
%It would be worth pointing out that HASSLE and SUMA are not the same method.HASSLE is the framework with the dual models to deal with annotation heterogeneity.It only requires that the dual models are trained separately on different types of data and should learn from each other to maximize the advantage of the framework.The detailed mutual learning strategy is specifically proposed as SUMA for the HASSLE framework, but it is not the only solution.
\begin{table*}[!t]
	\centering
	\begin{tabular}{@{}llccccccccc@{}}
		\toprule
		\multirow{2}{*}{\textbf{Family}}                                          & \multirow{2}{*}{\textbf{Method}} & \multicolumn{3}{c}{\textbf{FMIST}} & \multicolumn{3}{c}{\textbf{CIFAR-10}} & \multicolumn{3}{c}{\textbf{CIFAR-100}} \\ \cmidrule(l){3-11} 
		&                                  & 5\%           & 10\%        & 20\%       & 5\%          & 10\%         & 20\%         & 5\%             & 10\%        & 20\%        \\ \midrule
		\multirow{2}{*}{\begin{tabular}[c]{@{}l@{}}Supervised\\FL methods\end{tabular}} 
		&FedAvg     & 69.09       
		& 78.25      
		& 82.95
		& 50.23     & 57.85  & 77.65     & 17.36      & 28.43     &  39.52     \\
		&	FedProx     & 69.71       
		& 79.14     & 83.17     & 50.76   & 58.43   & 78.22     & 17.51      & 28.76      &  39.89       \\\midrule
		\multirow{4}{*}{\begin{tabular}[c]{@{}l@{}}Semi-supervised\\ FL methods\end{tabular}}          
		&FedAvg+MixMatch    & 70.45      & 79.67      & 83.45      & 51.85      & 59.18 & 66.28      & 17.96      & 25.11      & 40.53      \\
		&FedAvg+FixMatch    & 73.09      & 80.83      & 83.93     & 52.18      & 61.06   & 78.34      & 17.96      & 29.56      &  41.94     \\
		&FedMatch     & 75.54     & 82.87      & 85.05      & 56.67      & 63.38    & 79.73     & 21.27      & 32.58      & 42.72      \\
		&RSCFed     & 75.26      & 82.54      & 85.14     & 55.71      & 63.76   &  79.33     & 19.41      & 31.43      & 42.21     \\\midrule  
		\multirow{3}{*}{Proposed methods}           
		&HASSLE-SM    & 77.83      & 83.22      & 86.15      & 60.47      &66.34   & 80.78      & 25.52      & 34.11      & 43.05      \\
		&HASSLE-UM    & 77.98      & 83.91      & 86.34      & 60.98      &67.14   & 81.16     & 26.15      & 34.25      & 43.95      \\
		&HASSLE-EM    & \textbf{78.04}      & \textbf{84.04}      & \textbf{86.67}      & \textbf{61.27}      &\textbf{67.79}   & \textbf{81.52}     & \textbf{26.58}      & \textbf{34.33}      & \textbf{44.24}      \\ \bottomrule                                	
	\end{tabular}
	\caption{Top-1 test accuracy ($\%$) achieved by compared methods and HASSLE on three datasets with different percentages of labeled data.}
	\label{t1}
	\vspace{-10pt}
\end{table*} 
\section{Experiments}
\subsection{Experimental Setup}\label{exp_setup}
\paragraph{Datasets and Models.}
%数据集
\vspace{-5pt}
We evaluate our proposed framework HASSLE and learning strategy SUMA on typical classification benchmarks, including FMNIST \cite{xiao2017fashion}, CIFAR-10, and CIFAR-100 \cite{2009Learning}.
% 网络
We use LeNet \cite{lecun1998gradient} for FMNIST, and ResNet-8 \cite{he2016deep} for CIFAR-10/100 as the base encoder for the dual models. The residual models are more compact that use the same structure of the dual models with only 1/4 of the number of channels. We implement all compared FL methods with the same model for a fair comparison. 

\vspace{-10pt}
\paragraph{Data Distribution with Annotation and Data Heterogeneity.}
Following the previous study \cite{NEURIPS2020_18df51b9}, we generate the heterogeneous data among clients by Dirichlet distribution parameterized by $\alpha$. We set the value of $\alpha$ at 0.1. More results with different $\alpha$ are shown in \cref{tabel_non_iid} in Appendix C.1. 
With heterogeneous annotation, there are at most three kinds of clients: fully-labeled clients, partially-labeled clients, and unlabeled clients. 
The annotation distribution varies according to the percentage of labeled data (see the experiments in \cref{section_comparison}) and the type of annotation heterogeneity (see the third part of \cref{section_ah}).
We set the number of total clients at 20, which is a typical experimental setting in FL \cite{tan2022fedproto} and can cover all kinds of clients and types annotation heterogeneity settings. 
By default, we set one fully-labeled client, nine partially-labeled clients, and ten unlabeled clients in our experiments. In each round, 40\% of all clients are randomly selected.
The annotation distribution and data distribution of CIFAR-10 is shown in \cref{fig_data_dis} in Appendix B.1.
\vspace{-10pt}
\paragraph{Baselines and Training Details.}
We compare the proposed method with several FL methods: 
1) FedAvg \cite{mcmahan2017communication} and 2) FedProx \cite{MLSYS2020_38af8613} are baselines for supervised FL with only labeled data. 
3) FedAvg+MixMatch \cite{berthelot2019mixmatch} and 4) FedAvg+FixMatch \cite{sohn2020fixmatch} apply supervised FL on fully-labeled clients and semi-supervised learning by MixMatch/FixMatch on partially-labeled clients, then local models are aggregated on the server by FedAvg. 
5) FedMatch \cite{jeong2020federated} and 6) RSCFed \cite{liang2022rscfed} are the state-of-the-art FSSL methods.
For the proposed HASSLE, we evaluate the performance of both the supervised model (HASSLE-SM) and the unsupervised model (HASSLE-UM). Both of them use the output in tandem with the residual models by summing the logits up. In addition, we also evaluate the ensemble model (HASSLE-EM) of the supervised and unsupervised models, which is calculated by the mean outputs of HASSLE-SM and HASSLE-UM.
All experiments are repeated with 3 different random seeds. By default, we run 200 communication rounds for all methods. For local training, the batch size is set at 128. We use SGD with a learning rate of 0.1 and momentum of 0.9 as the optimizer. 
%\paragraph{Implementation.}
%%编程语言与参数
\vspace{-2pt}
\subsection{Comparative Results}
\vspace{-2pt}
\paragraph{Comparison with the State-of-the-art Methods.}\label{section_comparison}
% 如何切分数据，在5%的时候，只有一个client是labeled data，其他的都是unlabeled client
Following the typical setting of SSL, the total amount of available labeled data is limited. 
We vary the percentages of labeled data in the set \{5\%, 10\%, 20\%\}. 
%For the case of 5\%, there is only one fully-labeled client, and all the others are unlabeled clients. 
%For other cases, all three kinds of clients participate.
The corresponding detailed client distributions are depicted in \cref{table_annotation} in Appendix B.2. \cref{t1} shows the results of HASSLE compared with other methods on three datasets with different percentages of labeled data. 
It shows that all HASSLE-based models achieve the highest test accuracy.
% 与FedAvg相比
Compared with the baseline FedAvg, the performance gain of HASSLE-EM is the highest because FedAvg only utilizes the labeled data, which leads to a waste of unlabeled data. 
%When the percentage of labeled data is only 5\% (around 8.95\%, 11.04\%, and 9.22\% for three benchmark datasets, respectively) 
% 与其他半监督方法相比
The other semi-supervised FL methods (e.g., FedAvg+FixMatch and FedMatch) perform well in some cases compared with FedAvg. However, there is still a performance gap compared with HASSLE because they only utilize fully-labeled and partially-labeled clients and ignore unlabeled clients. RSCFed takes the unlabeled clients into account. However, it does not fully exploit the information of partially-labeled clients.
% 我们的
%On the contrary, the proposed HASSLE can be simply applied to arbitrarily-labeled clients to fully utilize different types of data.
On the contrary, the proposed HASSLE can be simply applied to arbitrarily-labeled clients to fully extract the knowledge of both labeled and unlabeled data, which are beyond the ability of existing FL methods.

%不同比例的结论
\begin{figure*}[]
	\centering
	\begin{minipage}[t]{0.305\linewidth}
		\centering
		\includegraphics[width=\linewidth]{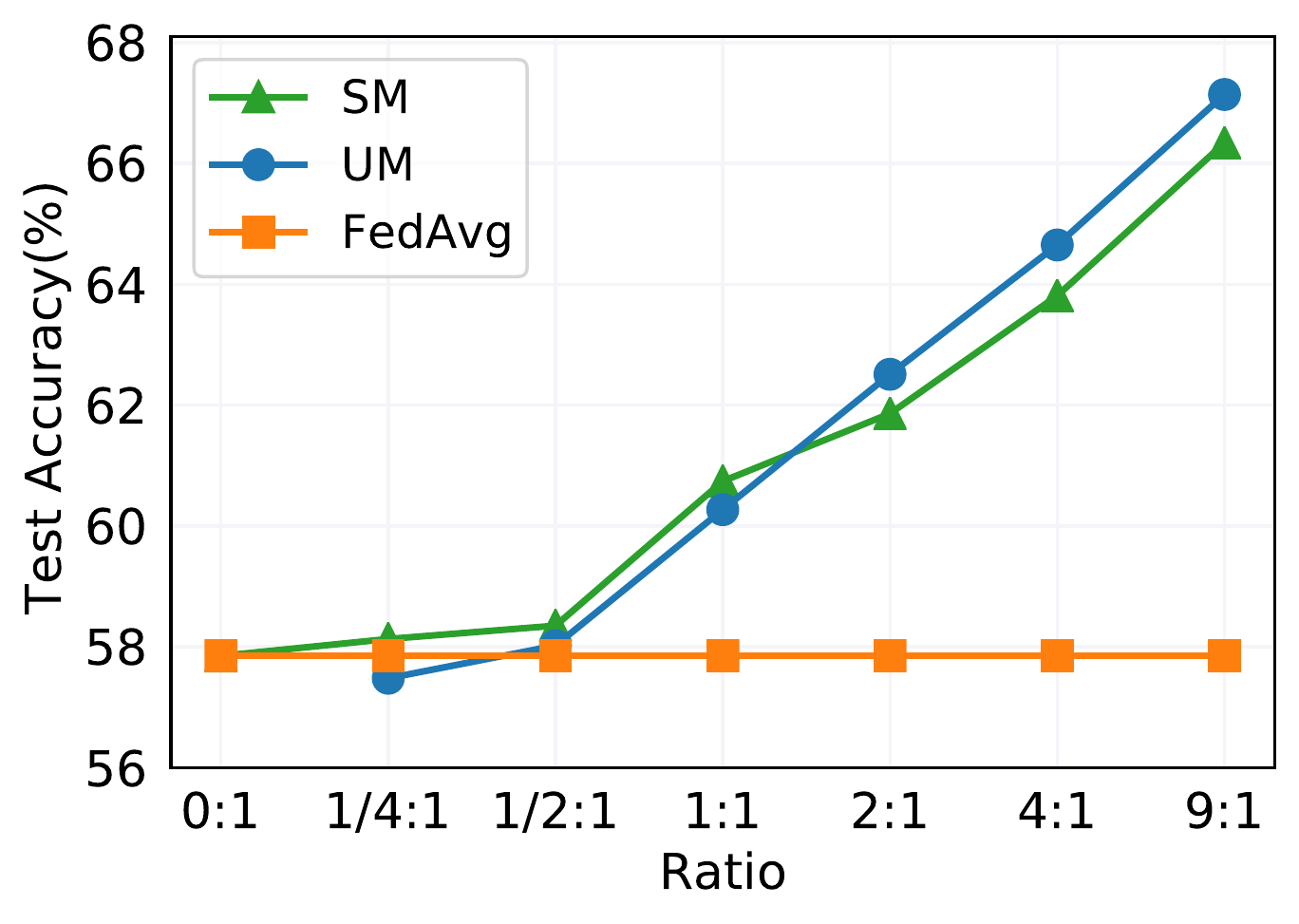}
		\caption{The performance of HASSLE with the different ratios between $\DD_{\UU}$ and $\DD_{\DL}$ on CIFAR-10.}
		\label{num_unlabeled}
		\vspace{-7pt}
	\end{minipage}
	\hspace{0.05in} % 两图片之间的距离
	\begin{minipage}[t]{0.31\linewidth}
		\centering
		\includegraphics[width=\linewidth]{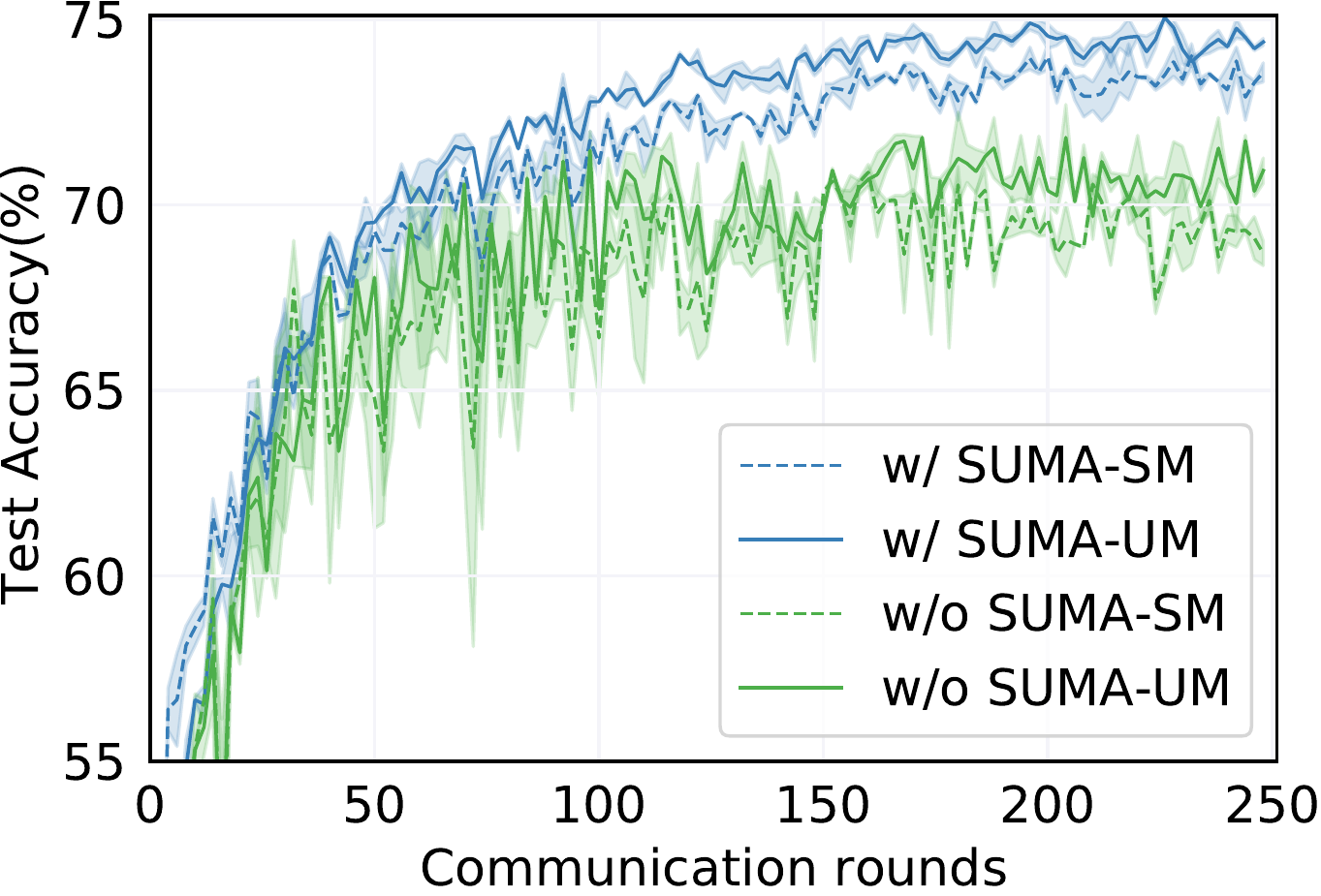}
		\caption{The performance in each round of SM and UM in HASSLE with or without SUMA on CIFAR-10.}
		\label{fig_mutual_learning}
		\vspace{-7pt}
	\end{minipage}
	\hspace{0.08in} % 两图片之间的距离
	\begin{minipage}[t]{0.30\linewidth}
		\centering
		\includegraphics[width=\linewidth]{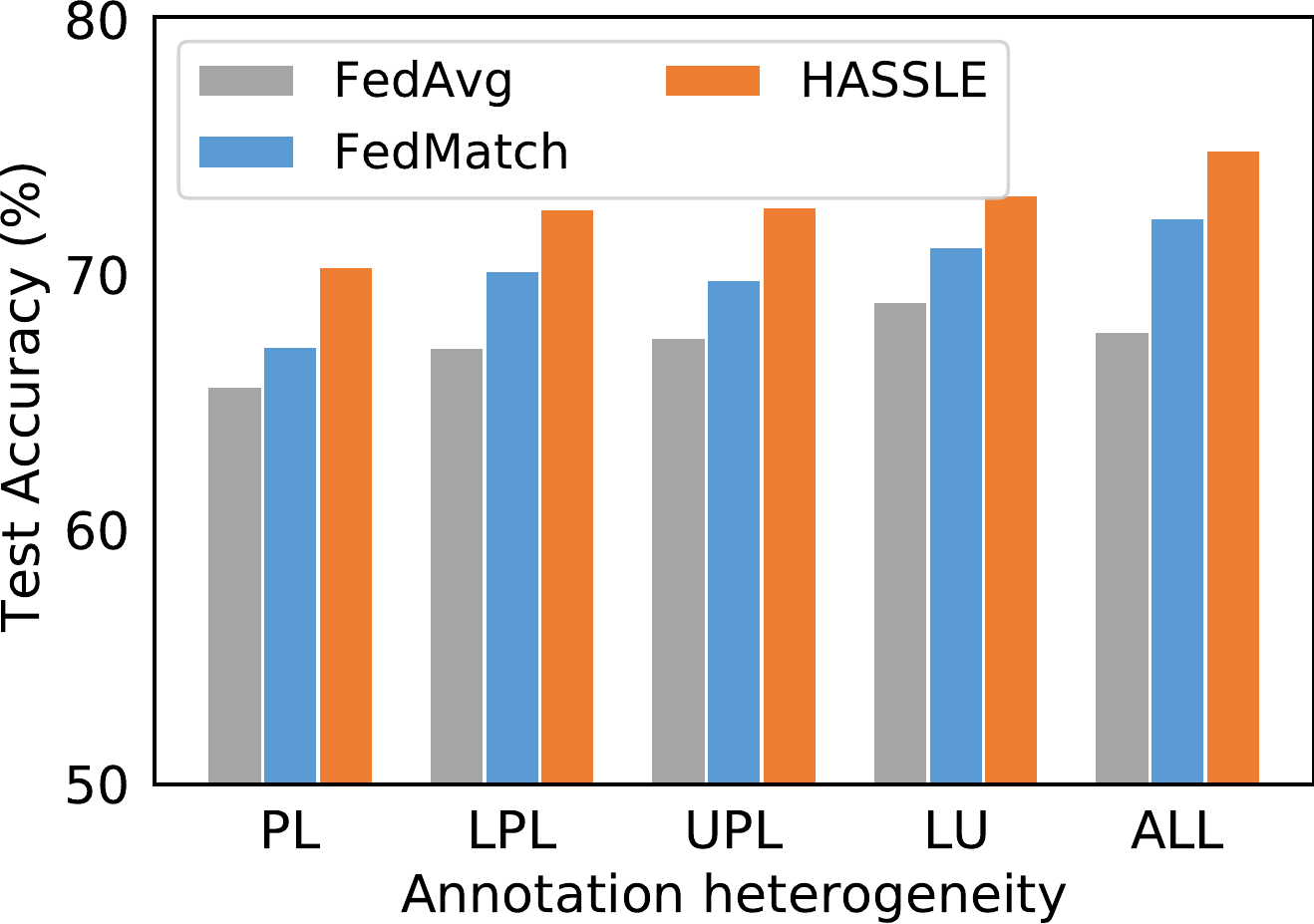}
		\caption{The performance of compared methods in five heterogeneous annotation settings on CIFAR-10.}
		\label{fig_annotation}
		\vspace{-7pt}
	\end{minipage}
\end{figure*} 
\vspace{-12pt}
%三种方法对比
\paragraph{Comparing Three HASSLE-based Models.}
As shown in \cref{t1}, HASSLE-EM achieves the best performance, which verifies that the dual models are still slightly diverse, making their ensemble model perform slightly better. Moreover, it can be observed that HASSLE-UM performs better than HASSLE-SM in all cases. A plausible reason is that the number of unlabeled data for training HASSLE-UM is large, although its pseudo-labels may not be accurate. 

To further evaluate it, we conduct an experiment to evaluate the influence of the number of unlabeled data. We fix the number of labeled data and increase the number of unlabeled data, shown in \cref{num_unlabeled}. It shows that when the number of unlabeled data is small (e.g., ratio $\leq$ 1), the performance of UM is worse than SM. 
%When the ratio $=$ 1, UM still performs worse than SM because the quality of pseudo-labels cannot be comparable with the ground-truth labels. 
However, when the number of unlabeled data continues to increase (e.g., ratio $\geq$ 2), UM starts outperforming SM, which further verifies our claim that the large number of unlabeled data is the reason why UM performs better than SM. Moreover, as the number of unlabeled data increases, SM also becomes better compared with FedAvg because the dual models can fully learn from each other, which also evaluates the effectiveness and superiority of the HASSLE framework.

\subsection{Model Evaluation}\label{model_evaluation}
\vspace{-5pt}
\paragraph{Ablation Study on Mutual Alignment Strategies.}
To evaluate the effectiveness of the key modules used in SUMA, we conduct an ablation study on global residual alignment (GRA) and model proximity alignment (MPA) with three HASSLE models, as shown in \cref{ablation_table}.
We compare our full model with the cases in which GRA or MPA is absent.
When both of them are not used, the resulting dual models are trained separately only with the pseudo-label assignment. 
%From the results, it can be observed that both GRA and MPA improve the dual models. 
Comparing the cases of solely using MPA or GRA, MPA improves the dual model more because it directly makes the dual models closer in the parameter space.
GRA constantly improves the dual model whether MPA is present or not, which shows the effectiveness of the residual models.
Therefore, both alignment strategies improve the dual models from different perspectives, which is consistent with our analysis in \cref{mutual_learning}.
%With both GRA and MPA present, the accuracy of UM is only slightly higher than SM with a gap of 0.65\%, which indicates that the knowledge of the dual models is better transferred to each other, which makes their prediction accuracy closer.
%%%%
%non_iid程度的图，放在一起讲
\vspace{-10pt}
\paragraph{Effectiveness of Mutual Learning.} \label{sec_mutal_learning}
%CIFAR10-10%-1
% 画一个图即可（曲后面几个round，差别较大）
%%从图中说明的信息：1.双模型都比单独学要好；2.开始U没有L好，因为U依靠L打的标签
In this experiment, we show how the dual models gradually learn from each other by SUMA.
We compare the dual models SM and UM in HASSLE with or without SUMA in \cref{fig_mutual_learning}.
In the early training rounds, no matter with or without SUMA, UM performs worse than SM because SM cannot produce high-quality pseudo-labels for the unlabeled data to update UM. 
However, as the number of communication rounds increases, SM gradually performs better and can produce relatively high-quality pseudo-labels for a large amount of unlabeled data, which helps UM perform better.
In addition, it can also be verified that SM and UM with SUMA fully learn from each other in each round because the gap in their accuracy is always small as learning goes on (0.94\% in the last round), compared with the large gap between SM and UM without SUMA (2.29\% in the last round). The training loss in each round is shown in \cref{fig_training_loss} in Appendix C.3.
\vspace{-12pt}
\paragraph{Variations of Annotation Heterogeneity.}\label{section_ah}
% 五种setting———柱状图
%纯半监督； 2个有标签和18个无标签；纯半监督和无监督（10 + 10）；1+9+10；有监督+半监督
In this experiment, we design five different heterogeneous annotation settings to validate the robustness of SUMA: 1) PL: only partially-labeled clients; 2) LPL: both fully-labeled and partially-labeled clients; 3) UPL: both unlabeled and partially-labeled clients; 4) LU: fully-labeled and unlabeled clients; 5) ALL: fully-labeled clients, partially-labeled clients and unlabeled clients.
The five settings of annotation distribution are shown in \cref{table_annotation_ah} in Appendix B.2.
Note that we do not test the cases of all fully-labeled clients or all unlabeled clients because they do not follow the general setting of FSSL. \cref{fig_annotation} reports the results, which further verify that the proposed HASSLE can properly address the problem of FSSL with all kinds of annotation heterogeneity.
% of FedAvg, FedMatch and HASSLE

\vspace{-12pt}
\paragraph{Ablation Study on Structure of Residual Models.}\label{residual_models}
We conduct an ablation study on the structure of the residual models, shown in \cref{structure_residual}. Even using very compact residual models (e.g., 1/16 or 2/16 of the number of channels of the dual models) can still achieve performance gain (around 0.5\% improvement from None to 1/16 and 1/16 to 2/16 for all SM, UM, and EM). The reason is that they act as auxiliary models in the HASSLE and only need to learn the complementary knowledge between the dual models, as discussed in \cref{mutual_learning}. In addition, it can also be observed that the performance gain becomes slight as the size of the residual model continues to increase (only around 0.1\% from 4/16 to 16/16). Therefore, larger residual models can hardly provide more assistance because the complementary knowledge between the dual models is limited to some content.
%information to assist the learning of the dual models, because the complementary knowledge between the dual models is limited to some content.
%\paragraph{Influence of the Degree of data heterogeneity.}
%Figure \ref{fig_non_iid} further shows the test accuracy of four methods under the different degrees of data heterogeneity. It can be observed that the performance of all methods drops as the degree of data heterogeneity increases. However, the performances of the compared methods drop more severely than HASSLE when $\alpha$ decreases from 1.0 to 0.1. It validates that HASSLE also takes the heterogeneous data distribution into account, which is discussed in Section \ref{L2}. More detailed results are depicted in Table \ref{tabel_non_iid} in Appendix \ref{appendix_non_iid}.

% 放在消融里面一起讲吧，就不要放这里了
%\subsubsection{Residual and L2}
% 当non-iid小的时候，residual没啥用，此时主要是L2有作用
%\paragraph{Hyperparameter Stability.}
%We investigate the impact of the hyperparameter $\lambda$ of residual model update in Equation (\ref{residual_model_update}) and the hyperparameter $\gamma$ of model proximity alignment in Equation (\ref{L2_optimization_1})-(\ref{L2_optimization_2}). Figure \ref{fig_l2_weight} and \ref{fig_kl_weight} in Appendix \ref{hyperparameter} shows that HASSLE is robust to most of $\lambda$ and $\gamma$ values.
\begin{table}[!t]
	\centering 
	\begin{tabular}{@{}c|c|ccc@{}}
		\toprule
		GRA   & MPA & SM & UM & EM \\ \midrule
		\xmark & \xmark & 57.85 & 61.07 & 61.86 \\
		\xmark & \cmark & 65.04 & 65.98 & 66.47 \\
		\cmark & \xmark & 62.11 & 64.09 & 64.54 \\
		\cmark & \cmark & \textbf{66.34} & \textbf{67.14} &\textbf{67.79} \\ \bottomrule
	\end{tabular}
	\caption{Ablation study of SUMA on CIFAR-10.}
	\label{ablation_table} 
	\vspace{-5pt}
\end{table}
\begin{table}[!t]
	\centering
	\begin{tabular}{@{}cccc@{}}
		\toprule
		\textbf{Structure} & \textbf{SM} & \textbf{UM} & \textbf{EM} \\ \midrule
		None                              & 65.04       & 65.98       & 66.47       \\ 
		ResNet-8 (1/16)                   & 65.57       & 66.47       & 67.13       \\ 
		ResNet-8 (2/16)                   & 65.86       & 66.73       & 67.44       \\ 
		ResNet-8 (4/16)                   & 66.34       & 67.14       & 67.79       \\ 
		ResNet-8 (16/16)                  & 66.46       & 67.28       & 67.85       \\ \bottomrule
	\end{tabular}
	\caption{Ablation study on the structure of the residual models on CIFAR-10.}
	\label{structure_residual}
	\vspace{-10pt}
\end{table}
\vspace{-5pt}
\section{Conclusion}\label{conclusion}
In this paper, we study a more general and realistic problem setup of FSSL with \emph{annotation heterogeneity}, where each client can hold an arbitrary percentage of labeled data.
Accordingly, we propose the HASSLE framework with the dual models separately trained on labeled and unlabeled data, which can be simply applied to arbitrary-labeled clients. Subsequently, we propose a mutual learning strategy SUMA for the dual models in HASSLE.
In SUMA, we adopt two additional residual models to learn the difference between the dual models to complement them from each other with model proximity constraints. Experiments verify that HASSLE with SUMA can well solve the problem of annotation heterogeneity by effectively extracting the knowledge of arbitrarily-labeled clients.
%learning from arbitrarily-labeled clients effectively. %In our future work, we plan to extend our framework to FL scenarios with more generalized data distribution, not limited to FSSL.

%%%%%%%%% REFERENCES
{\small
\bibliographystyle{ieee_fullname}
\bibliography{fl_ref}
}
%%%%%%%%%%%%%%%%%%%%%%%%%%%%%%%%%%%%%%%%%%%%%%%%%%%%%%%%%%%%
\newpage
%\title{Federated Semi-Supervised Learning with Annotation Heterogeneity}
\maketitle
\appendix
\onecolumn
\begin{center} 
	{\centering\section*{Appendix}}
\end{center}

\section{Pseudo-code of the Proposed Method}\label{pseudocode}
\cref{alg:algorithm} details the training procedure of the proposed SUMA in the HASSLE framework. In general, SUMA still follows the learning framework of FedAvg only with different local updating schemes. There are four steps in each round of our method. 
First, the server sends four models to clients, including the dual models ($\bw_{\MS}^{(t)}$ and $\bw_{\UU}^{(t)}$), and the two compact residual models $\br_{\UU-\MS}^{(t)}$ and $\br_{\MS-\UU}^{(t)}$).
Second, three kinds of clients update the received models in different ways. Specifically, for clients with labeled data (e.g., fully-labeled clients and partially-labeled clients), each of them updates the local supervised model $\bw_{\MS}^{(t+1)}$ by \cref{hassle_optimization}) and \cref{L2_optimization_1} and the residual model $\br_{\UU-\MS}^{(t+1)}$ by \cref{residual_model_update} on the local labeled data, respectively; for clients with unlabeled data (e.g., partially-labeled clients and unlabeled clients), the pseudo-labels for local unlabeled data are first assigned by \cref{pseudo_labeling_with_residual_model}. Then the local unsupervised model $\bw_{\UU}^{(t+1)}$ is updated by \cref{hassle_optimization} and \cref{L2_optimization_1} on the local unlabeled data with pseudo-labels, and the residual model $\br_{\MS-\UU}^{(t+1)}$ is updated by \cref{residual_model_update} on the local unlabeled data with pseudo-labels in a supervised learning manner, respectively.
Third, clients send their updated models to the server.
Last, the server aggregates local dual models to produce the global dual models ($\bw_{\MS}^{(t+1)}$ and $\bw_{\UU}^{(t+1)}$) by \cref{local_update} and aggregates local residual models to produce the global residual models ($\br_{\UU-\MS}^{(t+1)}$ and $\br_{\MS-\UU}^{(t+1)}$) by \cref{residual_model_aggregation} for the training of the next round. 

\begin{algorithm}[!h]
	\caption{The training process of SUMA in the HASSLE framework}
	\label{alg:algorithm}
	\LinesNumbered %要求显示行号
	\KwIn{Number of total clients $K$; Local dataset $\DD_k$ for $k=1,...,K$; Initialized models $\bw_\MS^{(0)}$, $\bw_\UU^{(0)}$, $\br_{\UU-\MS}^{(0)}$, and $\br_{\MS-\UU}^{(0)}$; Number of total rounds $T$; Number of selected clients $C$; Temperature coefficient $\tau$; Trade-off coefficient in residual model update $\lambda$; Trade-off coefficient in model proximity alignment $\gamma$; Learning rate $\eta$.}%输入参数
	%%%像是转置
	\KwOut{The global dual models $\bw_{\MS}^{(T)}$ and $\bw_{\UU}^{(T)}$, and the corresponding global residual models $\br_{\UU-\MS}^{(T)}$ and $\br_{\UU-\MS}^{(T)}$ on round $T$.}%输出
	\For{$t=1$ \KwTo $T$}{
		%\tcp{Server executes:}
		Randomly select a set of $C$ active clients $\mathcal{S}^t$;\\
		\tcp{Clients execute:}
		\For{$k\in \mathcal{S}^t$}{	
			\If{$n_{\DL,k}\ne0$}{
				Update the local supervised model $\bw_{\MS,k}^{(t+1)}$ by \cref{hassle_optimization} and \cref{L2_optimization_1};\\
				Update the local residual model $\br_{\UU-\MS,k}^{(t+1)}$ by \cref{residual_model_update};
			}
			\If{$n_{\UU,k}\ne0$}{
				Obtain the local pseudo-labeled subset $\hat{\DD}_{\UU,k}^{(t)}$ by \cref{pseudo_labeling_with_residual_model};\\
				Update the local unsupervised model $\bw_{\UU,k}^{(t+1)}$ by \cref{hassle_optimization} and \cref{L2_optimization_1};\tcp{Exchange $\UU$ and $\MS$}
				Update the local residual model $\br_{\MS-\UU,k}^{(t+1)}$ by \cref{residual_model_update}; \tcp{Exchange $\UU$ and $\MS$}
			}
			
		}
		\tcp{Server executes:}
		Aggregate local dual models to the global models $\bw_\MS^{(t+1)}$ and $\bw_\UU^{(t+1)}$ by \cref{local_update};\\
		Aggregate local residual models to the global models $\br_{\UU-\MS}^{(t+1)}$ and $\bw_{\MS-\UU}^{(t+1)}$ by \cref{residual_model_aggregation};\\
		Send the global dual models and global residual models to each client;\\
		%Send $\bw^{t+1}=\{\bu^{t+1}, \bv^{t+1}\}$ and $\widehat{\bw}^{t+1}=\{\bu^{t+1}, \widehat{\bv}^{t+1}\}$ to clients;\\
	}
\end{algorithm}

\section{Data and Annotation Distribution on Each Client}

\subsection{Data Distribution}\label{data_distribution}

\begin{figure*}[!t]
	\centering
	\begin{subfigure}[b]{0.32\linewidth}
		\includegraphics[width=\linewidth]{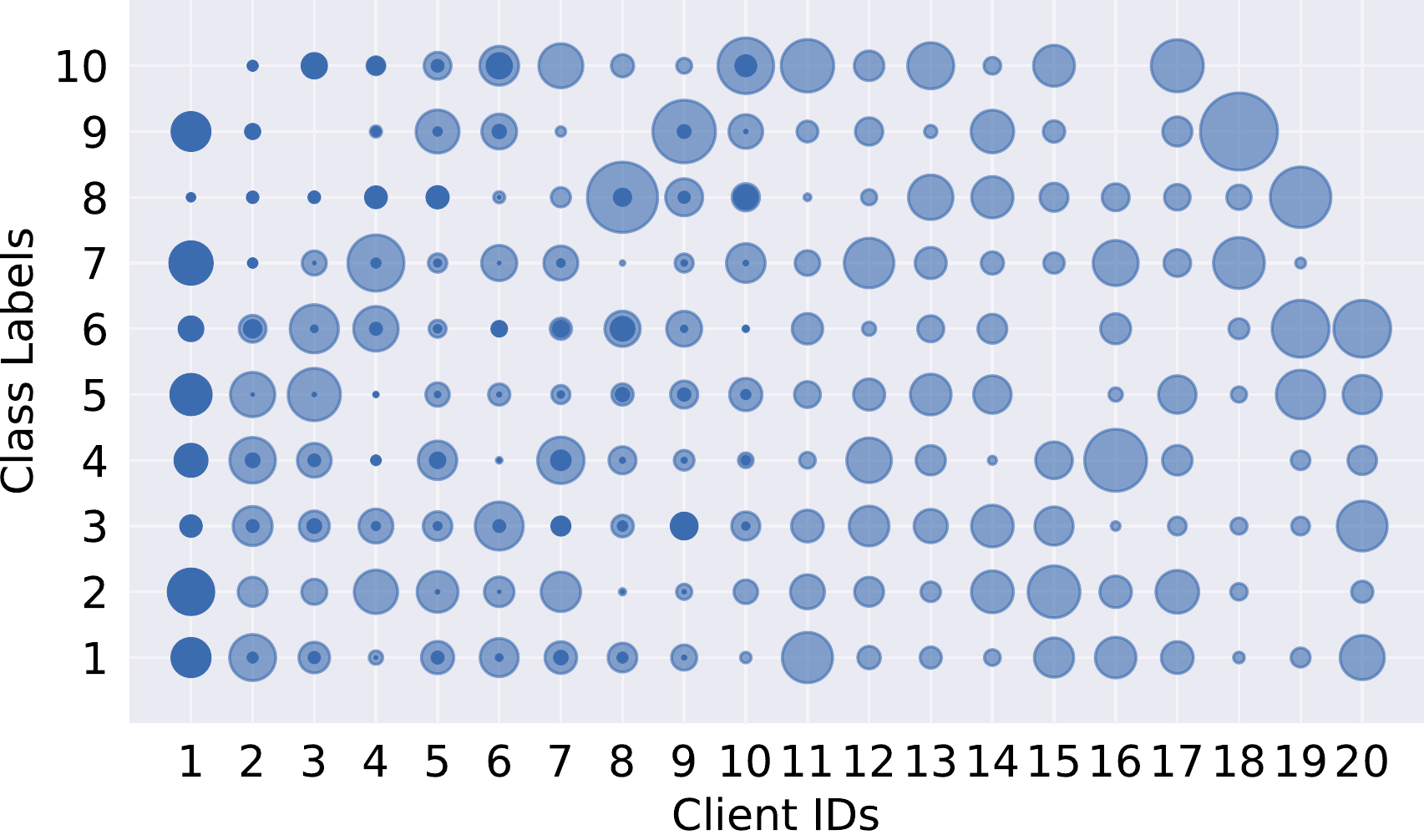}
		\subcaption{$\alpha=1$}
	\end{subfigure}
	\begin{subfigure}[b]{0.32\linewidth}
		\includegraphics[width=\linewidth]{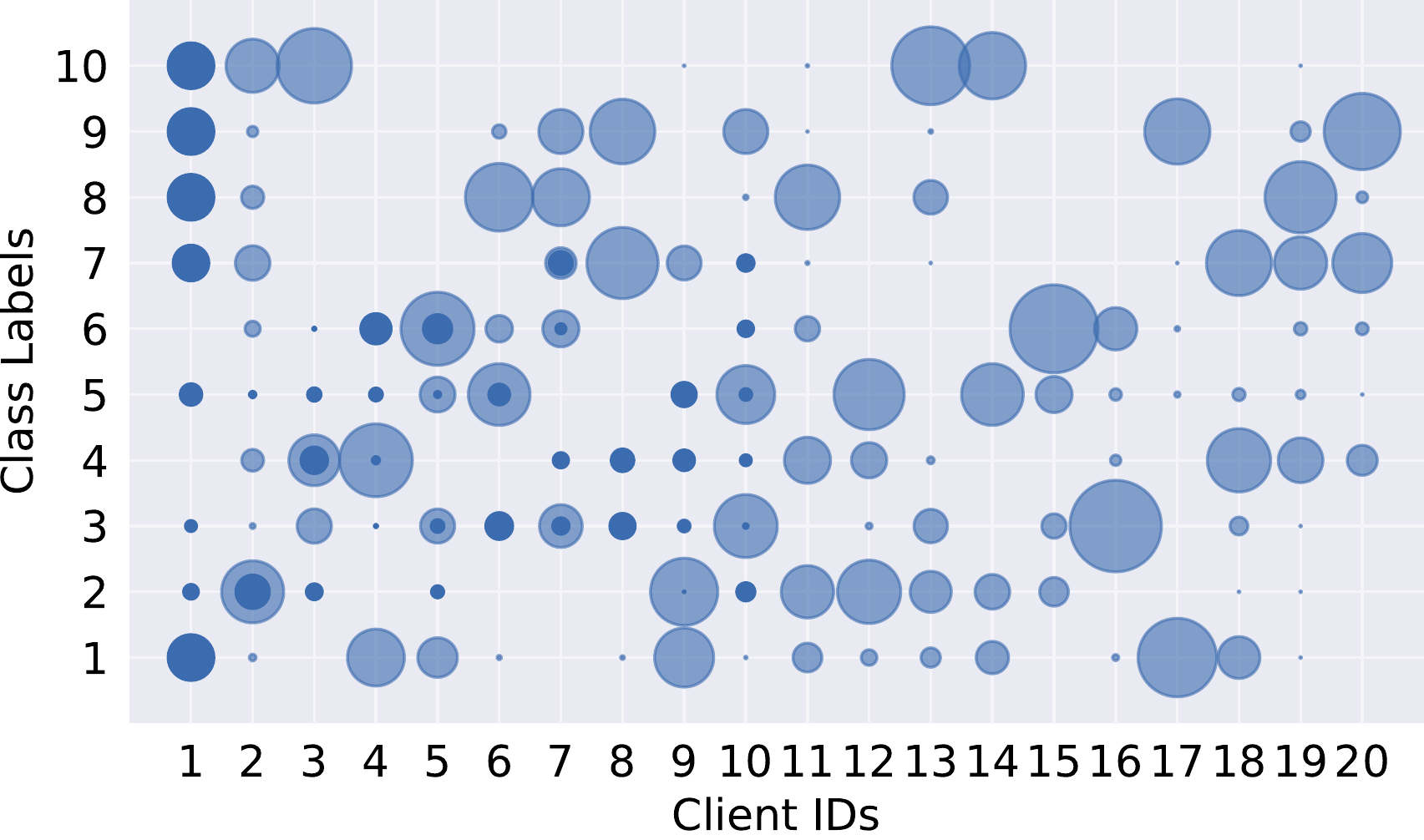}
		\subcaption{$\alpha=0.1$}
	\end{subfigure}
	\begin{subfigure}[b]{0.32\linewidth}
		\includegraphics[width=\linewidth]{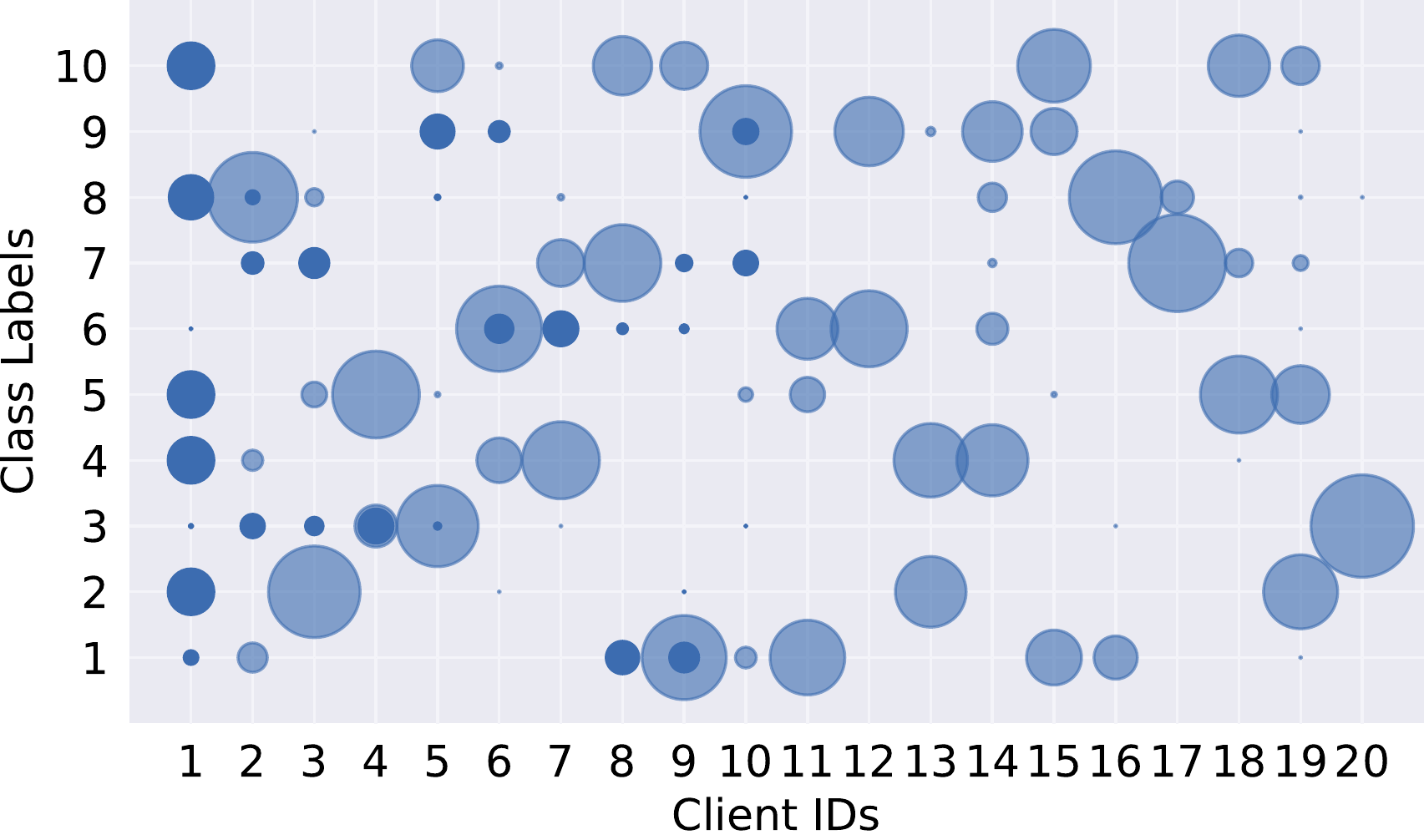}
		\subcaption{$\alpha=0.01$}
	\end{subfigure}\\
	\caption{Visualization of data distributions of CIFAR-10 with the different degrees of data heterogeneity. The transparent dots means unlabeled data, and the filled dots means labeled data. The size of each dot reflects the number of samples in each client.}
	\label{fig_data_dis}
\end{figure*}

Following the previous studies \cite{NEURIPS2020_18df51b9}, we use Dirichlet distribution to generate the heterogeneous data partition among clients. The value of $\alpha$ controls the degree of data heterogeneity. When $\alpha \rightarrow \infty$, all clients have identical distributions; When $\alpha \rightarrow 0$, all clients hold examples from only one random class.
To better understand the local data distribution for the datasets, we visualize the partition results of CIFAR-10 on $\alpha=\{0.01, 0.1, 1\}$ for 20 clients in \cref{fig_data_dis}.

\subsection{Annotation Distribution}\label{annotation_distribution}
\begin{table*}[!h]
	\centering
	%$\blacksquare$ and $\like{0}$ and $\like{4}$.
	\begin{tabular}{@{}lcccccccccccccccccccc@{}}
		\toprule
		& 1 & 2 & 3 & 4 & 5 & 6 & 7 & 8 & 9 & 10 & 11 & 12 & 13 & 14 & 15 & 16 & 17 & 18 & 19 & 20 \\ \midrule
		5\%  & $\like{4}$  & $\like{4}$  & $\like{0}$  & $\like{0}$  & $\like{0}$  & $\like{0}$  & $\like{0}$  & $\like{0}$  & $\like{0}$  &  $\like{0}$  & $\like{0}$   &  $\like{0}$  &  $\like{0}$  &  $\like{0}$  &  $\like{0}$  &  $\like{0}$  &  $\like{0}$  & $\like{0}$   &  $\like{0}$  & $\like{0}$   \\ \midrule
		10\% & $\like{10}$  & $\like{4}$  & $\like{4}$  & $\like{4}$  & $\like{4}$  &  $\like{4}$ &  $\like{4}$ & $\like{4}$  & $\like{4}$  & $\like{4}$   & $\like{0}$   & $\like{0}$   &  $\like{0}$  & $\like{0}$   & $\like{0}$   &  $\like{0}$  &  $\like{0}$  & $\like{0}$   &  $\like{0}$  & $\like{0}$   \\ \midrule
		20\% & $\like{10}$  & $\like{10}$  & $\like{4}$  &  $\like{4}$  & $\like{4}$   &  $\like{4}$  &  $\like{4}$  &  $\like{4}$  &  $\like{4}$  &  $\like{4}$   &  $\like{0}$  &  $\like{0}$  &  $\like{0}$  &  $\like{0}$  & $\like{0}$   &  $\like{0}$  & $\like{0}$   &  $\like{0}$  & $\like{0}$   & $\like{0}$   \\ \bottomrule
	\end{tabular}
	\caption{Heterogeneous annotation distribution of clients with different percentages of labeled data. $\like{10}$ means fully-labeled client; $\like{4}$ means partially-labeled client; $\like{0}$ means unlabeled client.}
	\label{table_annotation}
\end{table*}
\begin{table*}[!h]
	\centering
	\begin{tabular}{@{}lcccccccccccccccccccc@{}}
		\toprule
		& 1 & 2 & 3 & 4 & 5 & 6 & 7 & 8 & 9 & 10 & 11 & 12 & 13 & 14 & 15 & 16 & 17 & 18 & 19 & 20 \\ \midrule
		PL  & $\like{4}$  & $\like{4}$  & $\like{4}$  & $\like{4}$  & $\like{4}$  & $\like{4}$  & $\like{4}$  & $\like{4}$  & $\like{4}$  &  $\like{4}$  & $\like{4}$  &  $\like{4}$ & $\like{4}$  &  $\like{4}$  &  $\like{4}$  &  $\like{4}$  &  $\like{4}$  & $\like{4}$   &  $\like{4}$  & $\like{4}$   \\ \midrule
		LPL & $\like{10}$  & $\like{4}$  & $\like{4}$  & $\like{4}$  & $\like{4}$  &  $\like{4}$ &  $\like{4}$ & $\like{4}$  & $\like{4}$  & $\like{4}$   & $\like{4}$   & $\like{4}$   &  $\like{4}$  & $\like{4}$   &$\like{4}$   &  $\like{4}$  &  $\like{4}$ & $\like{4}$   &  $\like{4}$  & $\like{4}$   \\ \midrule
		UPL  & $\like{4}$  & $\like{4}$  & $\like{4}$  & $\like{4}$  & $\like{4}$  & $\like{4}$  &$\like{4}$  & $\like{4}$  & $\like{4}$  &  $\like{4}$  & $\like{0}$   &  $\like{0}$  &  $\like{0}$  &  $\like{0}$  &  $\like{0}$  &  $\like{0}$  &  $\like{0}$  & $\like{0}$   &  $\like{0}$  & $\like{0}$   \\ \midrule
		LU & $\like{10}$  & $\like{10}$  & $\like{0}$   & $\like{0}$   & $\like{0}$   &  $\like{0}$  &  $\like{0}$ & $\like{0}$   & $\like{0}$   & $\like{0}$    & $\like{0}$   & $\like{0}$   &  $\like{0}$  & $\like{0}$   & $\like{0}$   &  $\like{0}$  &  $\like{0}$  & $\like{0}$   &  $\like{0}$  & $\like{0}$   \\ \midrule
		ALL & $\like{10}$  & $\like{4}$  & $\like{4}$  &  $\like{4}$  & $\like{4}$   &  $\like{4}$  &  $\like{4}$  &  $\like{4}$  &  $\like{4}$  &  $\like{4}$   &  $\like{0}$  &  $\like{0}$  &  $\like{0}$  &  $\like{0}$  & $\like{0}$   &  $\like{0}$  & $\like{0}$   &  $\like{0}$  & $\like{0}$   & $\like{0}$   \\ \bottomrule
	\end{tabular}
	\caption{Clients distribution with different kinds of annotation heterogeneity. $\like{10}$ means fully-labeled client; $\like{4}$ means partially-labeled client; $\like{0}$ means unlabeled client.}
	\label{table_annotation_ah}
\end{table*}
With heterogeneous annotation, there are at most three kinds of clients: fully-labeled clients (marked as $\like{10}$), partially-labeled clients (marked as $\like{4}$), and unlabeled clients (marked as $\like{0}$).
The detailed client distributions with different percentages of labeled data in \cref{t1} are visualized in \cref{table_annotation}.
The detailed client distributions with different kinds of annotation heterogeneity in \cref{fig_annotation} are visualized in \cref{table_annotation_ah}.

\section{Additional Experimental Results}

\begin{wraptable}{R}{0.3\textwidth}\vspace{-10pt}
	\begin{tabular}{@{}c|ccc@{}}
	\toprule
	Threshold & SM & UM & EM \\ \midrule
	None      & 66.34     & 67.14     & 67.79     \\
	0.95      & 66.42     & 67.25     & 67.84     \\ \bottomrule
	\end{tabular}
	\caption{Top-1 test accuracy ($\%$) achieved by HASSLE with or without threshold}
	\label{threshold}
	\vspace{-25pt}
\end{wraptable}
%\color{red}
\subsection{Influence of the Threshold.}\label{sec_threshold}
%%改
For centralized SSL methods, such as FixMatch, threshold is used to select high-confidence unlabeled samples for model training with their pseudo-labels. However, in our setting, we show that the threshold does not play a major role. Although the pseudo-labels may not be accurate in the early stages of training, updating the unsupervised model on unlabeled data is not significantly affected because the proposed model proximity makes the local unsupervised model close to the global supervised model in the parameter space. In addition, as a complement to this unsupervised model, the unsupervised residual model can learn the knowledge from the supervised model to complement the unsupervised model. We also compared our method with the case where the threshold is set at 0.95, the same as FixMatch. The results are shown in \cref{threshold}. We can find that the performance of the two cases is similar, which verifies that the threshold is not necessary for our problem.

%\begin{table}[]
%\end{table}

\color{black}

\subsection{Comparative Results with Different Degrees of Data Heterogeneity} \label{appendix_non_iid}

\begin{table*}[!t]
	\centering
	\begin{tabular}{@{}llccccccccc@{}}
		\toprule
		\multirow{2}{*}{\textbf{Family}}                                          & \multirow{2}{*}{\textbf{Method}} & \multicolumn{3}{c}{\textbf{FMIST}} & \multicolumn{3}{c}{\textbf{CIFAR-10}} & \multicolumn{3}{c}{\textbf{CIFAR-100}} \\ \cmidrule(l){3-11} 
		&                                  & 0.01           & 0.1        & 1       & 0.01           & 0.1        & 1        & 0.01           & 0.1        & 1        \\ \midrule
		\multirow{2}{*}{\begin{tabular}[c]{@{}l@{}}Supervised\\FL methods\end{tabular}} 
		&FedAvg     & 62.14       
		& 78.25      
		& 82.51
		& 35.21     & 57.85  & 67.79     &   22.87    & 28.43     &  29.45     \\
		&FedProx     &  63.64      
		& 79.14     &   83.37   & 36.39   & 58.43   & 68.94     &   23.28    & 28.76      &  30.94    \\\midrule
		\multirow{4}{*}{\begin{tabular}[c]{@{}l@{}}Semi-supervised\\ FL methods\end{tabular}}          
		&FedAvg+MixMatch    &  65.57     & 79.67      &  83.41     &    45.95   & 59.18 & 69.42      & 25.85      & 25.11      & 31.34      \\
		&FedAvg+FixMatch    &  68.19     & 80.83      &  85.79    &   57.46    & 61.06   &    71.46   &   27.54    & 29.56      & 33.16     \\
		&FedMatch     & 70.28     & 82.87      &   86.51    &   60.17    & 63.38    &   72.29   &   29.95    & 32.58      &   34.53    \\
		&	RSCFed     &  69.16     & 82.54      &   86.12   &   60.77    & 63.76   &    72.38   &  29.36     & 31.43      &    34.09  \\ \midrule
	\multirow{3}{*}{Proposed method}  &	HASSLE-SM    &  72.43     & 83.22      &   87.18    &   62.19    &66.34   &   73.67    & 31.84      & 34.11      &   35.09    \\
	&	HASSLE-UM    &  73.54     & 83.91      &   87.34    &  63.17     &67.14   &  74.61    & 32.17      & 34.25      & 35.74      \\
	&	HASSLE-EM    & \textbf{73.79}      & \textbf{84.04}      & \textbf{87.51}      & \textbf{63.27}      &\textbf{67.79}   & \textbf{74.89}     & \textbf{32.56}      & \textbf{34.33}      & \textbf{35.95}      \\ \bottomrule
	\end{tabular}
	\caption{Top-1 test accuracy ($\%$) achieved by compared FL methods and HASSLE on three datasets with different degrees of data heterogeneity.}
	\label{tabel_non_iid}
\end{table*} 
The results on three typical image classification datasets with different $\alpha$ values are summarized in \cref{tabel_non_iid}. It can be observed that all HASSLE-based methods achieve the highest test accuracy with all different degrees of data heterogeneity. 

\begin{wrapfigure}{R}{0.3\textwidth}\vspace{-15pt}
	\centering
	\includegraphics[width=0.3\textwidth]{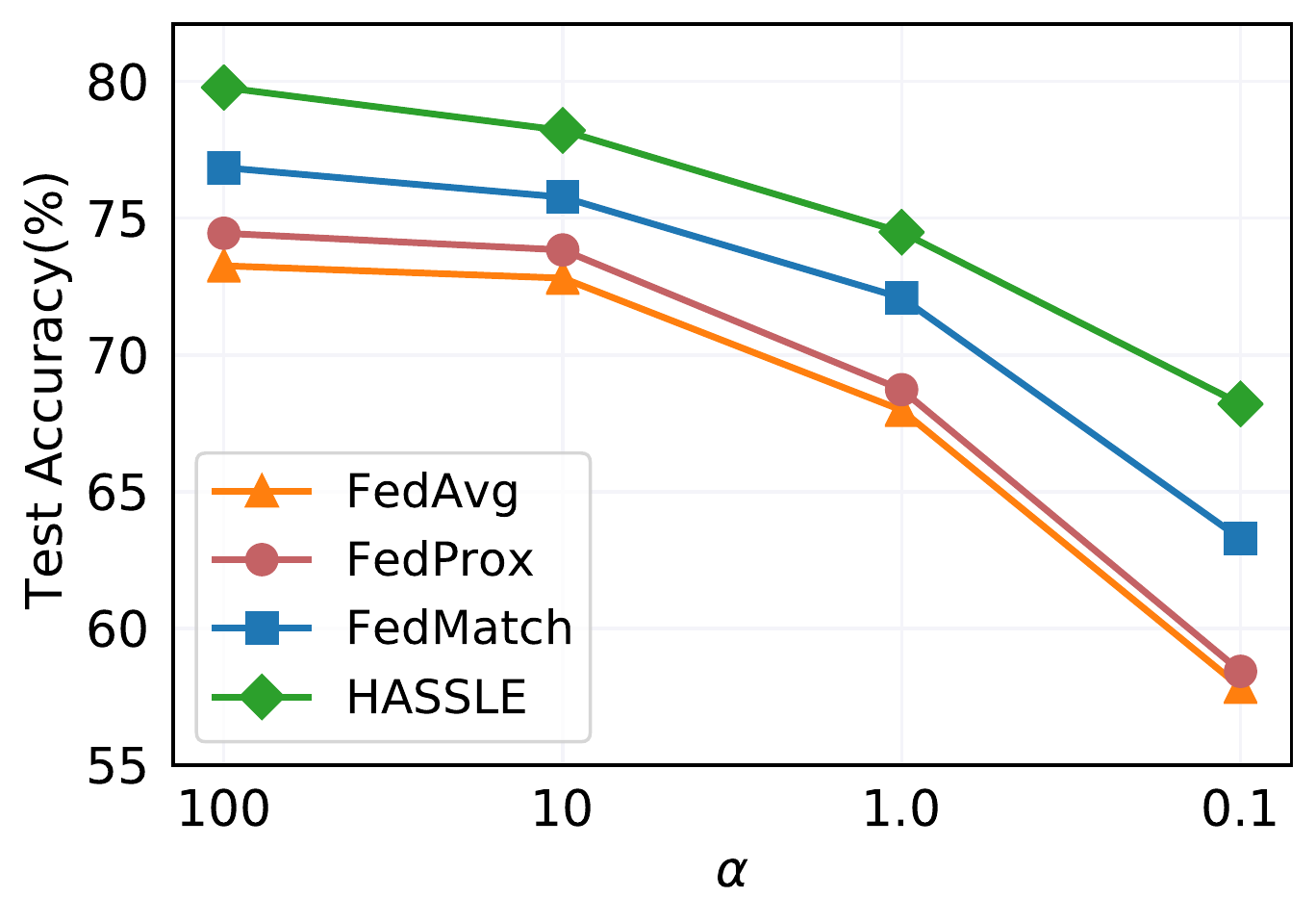}
	\caption{The performance of HASSLE with the different degrees of data heterogeneity on CIFAR-10.}
	\label{fig_non_iid}
	\vspace{-20pt}
\end{wrapfigure}

\subsection{Influence of the Degree of data heterogeneity.}
\cref{fig_non_iid} further shows the test accuracy of four methods under the different degrees of data heterogeneity. It can be observed that the performance of all methods drops as the degree of data heterogeneity increases. However, the performances of the compared methods drop more severely than HASSLE when $\alpha$ decreases from 1.0 to 0.1. It validates that HASSLE also takes the heterogeneous data distribution into account, which is discussed in \cref{L2}.

\subsection{Training Loss}\label{appendix_training_loss}
We show the training loss of the supervised and unsupervised models in \cref{fig_training_loss}. 
In the early rounds, the training loss of SM is smaller than UM because SM is not able to assign high-quality pseudo-labels at the beginning for the unlabeled data, which is consistent with our analysis in \cref{sec_mutal_learning}. 
However, as training goes on, the training losses of SM and UM become gradually stable and decrease to around 0.9.

\subsection{Hyperparameter Study}\label{hyperparameter}

\begin{figure*}[h]
	\centering
	\begin{minipage}[t]{0.31\linewidth}
		\centering
		\includegraphics[width=\linewidth]{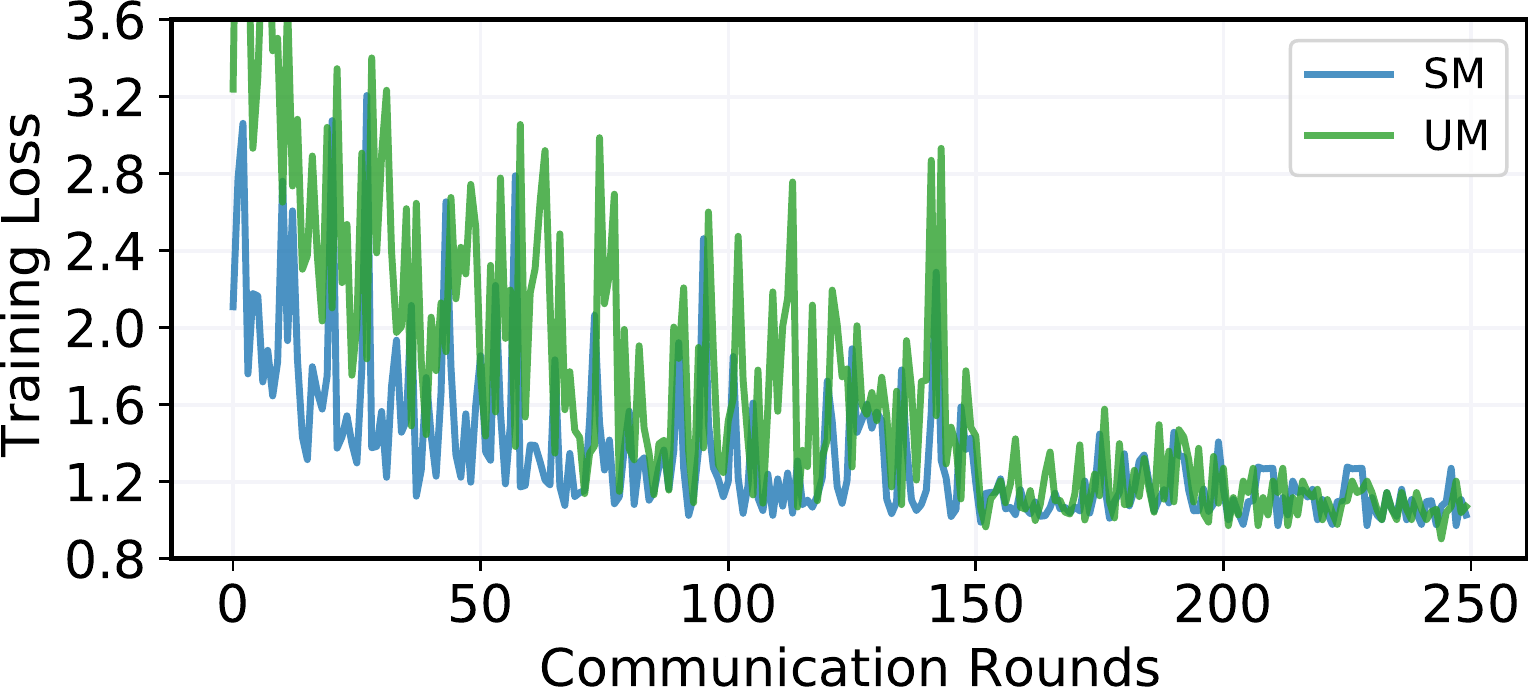}
		\caption{The training loss of SM and UM in each round.}
		\label{fig_training_loss}
	\end{minipage}
	\hspace{0.08in} % 两图片之间的距离
	\begin{minipage}[t]{0.31\linewidth}
		\centering
		\includegraphics[width=\linewidth]{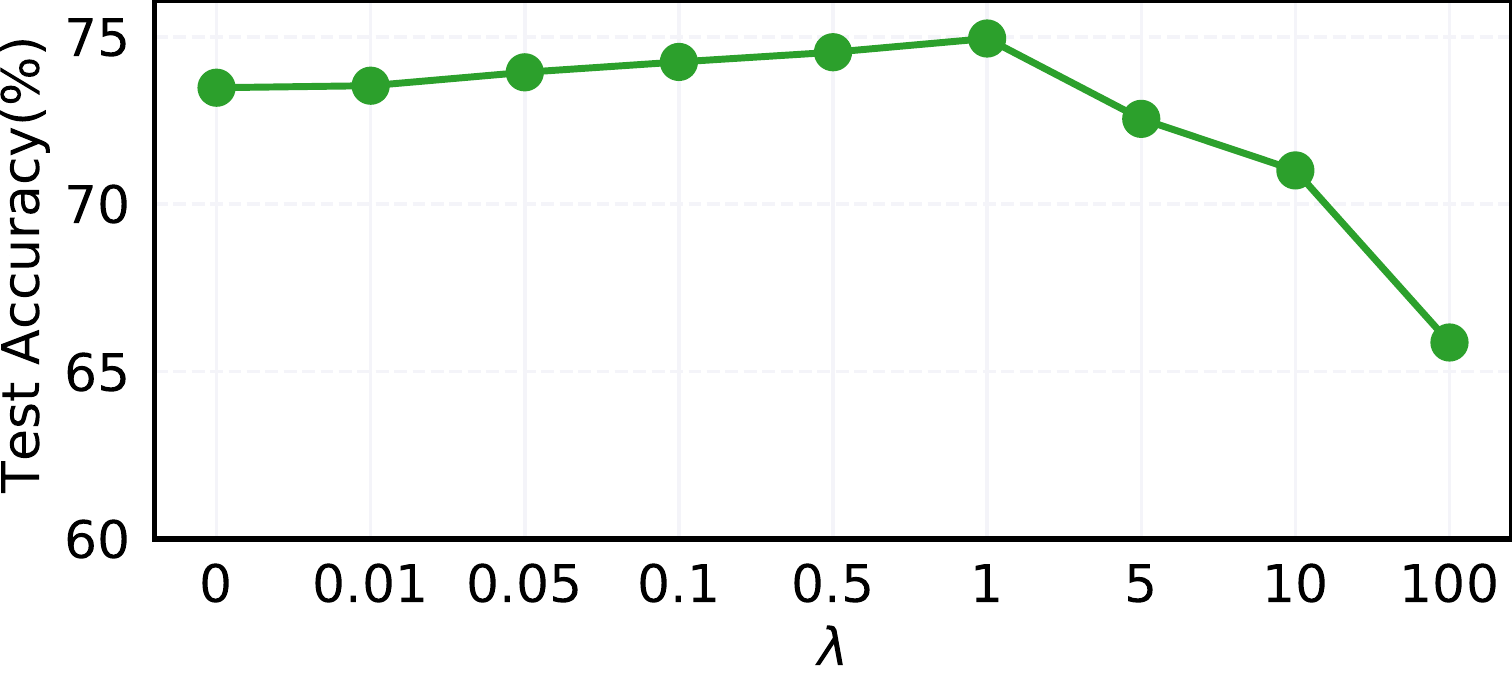}
		\caption{The performance of HASSLE with different values of $\lambda$ on CIFAR-10.}
		\label{fig_kl_weight}
	\end{minipage}
	\hspace{0.05in} % 两图片之间的距离
	\begin{minipage}[t]{0.31\linewidth}
		\centering
		\includegraphics[width=\linewidth]{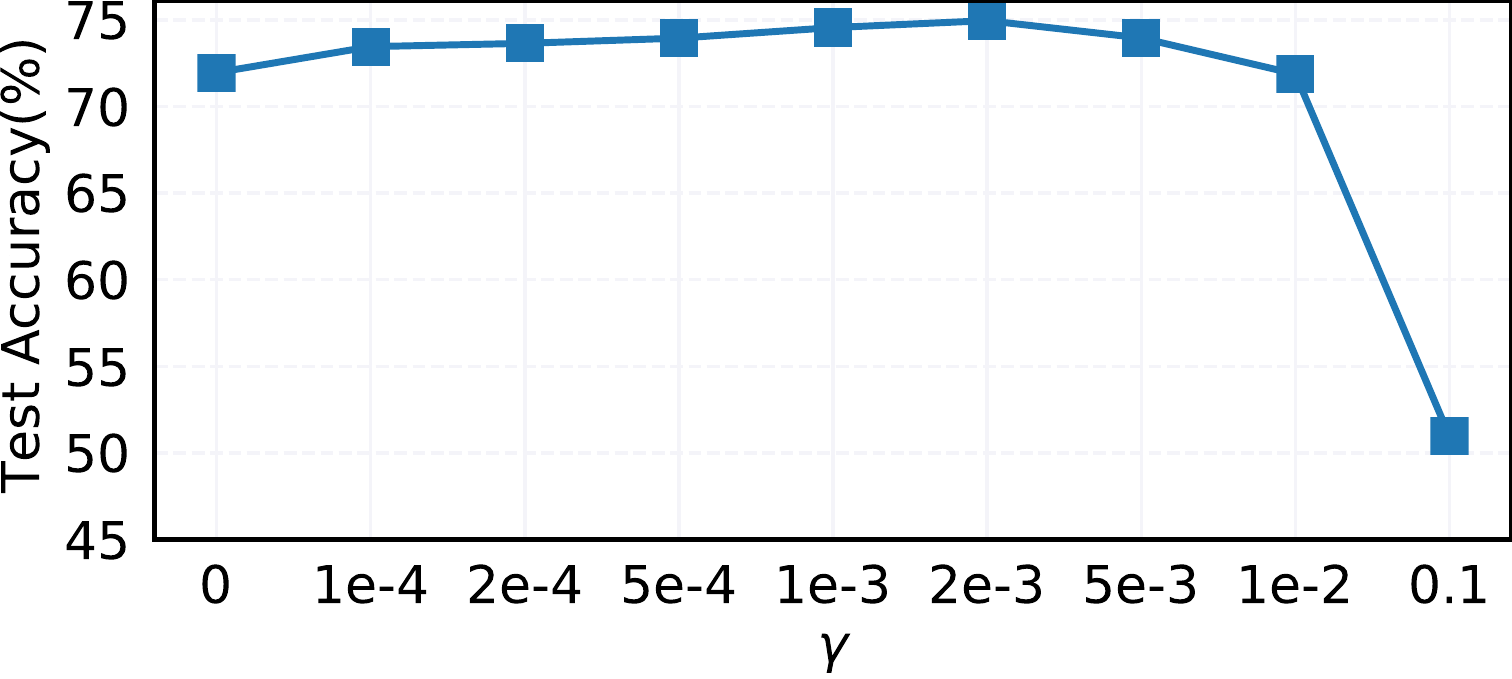}
		\caption{The performance of HASSLE with different values of $\gamma$ on CIFAR-10.}
		\label{fig_l2_weight}
	\end{minipage}
\end{figure*}

\paragraph{Hyperparameter $\lambda$ for Residual Model Update.}
We investigate the impact of the hyperparameter $\lambda$ for residual model update in \cref{residual_model_update}. \cref{fig_l2_weight} shows that the best performance is achieved when $\lambda=1$, i.e., the two losses are with the equal weight, which means both $L_{res}$ and $L_{CE}$ are necessary to update the residual models effectively.

\paragraph{Hyperparameter $\gamma$ for Model Proximity Alignment.}
We investigate the impact of the hyperparameter $\gamma$ of model proximity alignment in \cref{L2_optimization_1}. 
Figure \cref{fig_kl_weight} shows that HASSLE is robust to most of $\gamma$ values.
However, the performance drops when $\lambda=0.1$.
A plausible reason is that a relatively large value of $\lambda$ imposes the parameter exchange of the dual models without continually learning from the local data.

\end{document}